\documentclass[11pt]{article}

\usepackage[preprint]{acl}

\usepackage{times}
\usepackage{latexsym}

\usepackage[T1]{fontenc}

\usepackage[utf8]{inputenc}

\usepackage{microtype}
\newcommand{\cready}[1]{} 
\usepackage{amsmath} 

\usepackage[colorinlistoftodos]{todonotes}

\usepackage{enumitem}
  \setlist{itemsep=0.5pt, topsep=0pt,leftmargin=1.4em,labelsep=0.5em}

\usepackage[capitalize,nameinlink]{cleveref}
\Crefname{equation}{Eq.}{Eqs.}
\Crefname{figure}{Fig.}{Figs.}
\Crefname{tabular}{Tab.}{Tabs.}

\usepackage{etoolbox} 
\makeatletter
\pretocmd{\appendix}{%
  \@addtoreset{figure}{section}%
  \@addtoreset{table}{section}%
}{}{}
\makeatother
\usepackage{inconsolata}
\usepackage{amsfonts}
\usepackage{bbm}
\usepackage{booktabs}

\usepackage{graphicx}

\title{Will it Merge? \\ On The Causes of Model Mergeability}

\author{
 \textbf{Adir Rahamim\textsuperscript{1}},
 \textbf{Asaf Yehudai\textsuperscript{2,5}},
 \textbf{Boaz Carmeli\textsuperscript{1,2}},
 \textbf{Leshem Choshen\textsuperscript{3,4}},
\\
 \textbf{Yosi Mass\textsuperscript{2}},
 \textbf{Yonatan Belinkov\textsuperscript{1,6}}
\\
\\
 \textsuperscript{1}Technion - Israel Institute of Technology,
 \textsuperscript{2}IBM Research AI,
 \textsuperscript{3}MIT,
 \textsuperscript{4}MIT-IBM Watson AI Lab,
 \\
 \textsuperscript{5}Hebrew University of Jerusalem,
 \textsuperscript{6}Kempner Institute, Harvard University
\\
   \href{mailto:adir.rahamim@campus.technion.ac.il}{adir.rahamim@campus.technion.ac.il}
}

\begin{document}
\maketitle
\begin{abstract}
Model merging has emerged as a promising technique for combining multiple fine-tuned models into a single multitask model without retraining.
However, the factors that determine whether merging will succeed or fail remain poorly understood. In this work, we investigate why specific models are merged better than others. 
To do so, we propose a concrete, measurable definition of mergeability.  
We investigate several potential causes for high or low mergeability, highlighting the base model knowledge as a dominant factor: Models fine-tuned on instances that the base model knows better are more mergeable than models fine-tuned on instances that the base model struggles with.  
Based on our mergeability definition, we explore a simple weighted merging technique that better preserves weak knowledge in the base model.  
\end{abstract}

\section{Introduction}
Large pre-trained models are commonly fine-tuned on various downstream tasks to achieve better specialization on specific tasks. This task-specific model training has motivated model merging techniques \cite{matena2022merging, wortsman2022model, choshen2022fusing, ilharco2022editing, yu2024language, yadav2023ties, stoica2024model}, which aggregate the weights of multiple fine-tuned models into a single expert model that should perform well on all tasks. However, while algorithmic advances have improved merging algorithms, a fundamental question remains: What determines whether merging of a fine-tuned model will be successful? Answering this question might assist us in understanding how to obtain more mergeable models and inform better merging algorithms.



\begin{figure*}
    \centering
    \includegraphics[width=1\linewidth]{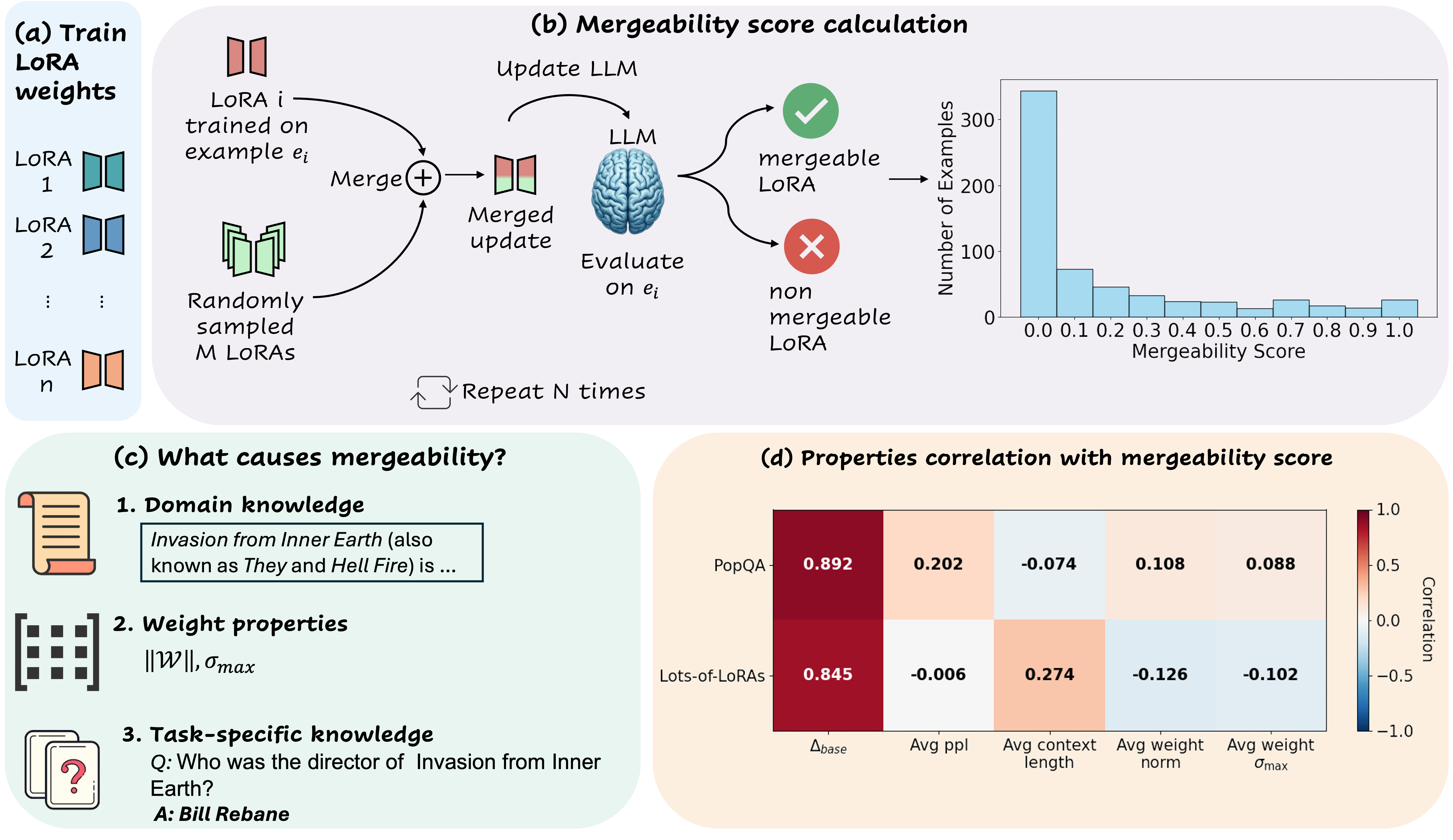}
    \caption{Our Experimental setup. The figure shows an example from PopQA, Lots-of-LoRAs experiments follow a similar process. (a) We train an expert model for each example, and verify the model's correctness on the example after training. (b) We calculate each example's mergeability score, as defined in \S\ref{sec:mergeability} and group examples by their mergeability score. (c) We investigate different traits that affect mergeability. Results are in \S\ref{sec:mergeability_causes}. (d) We examine the correlation between the mergeability score and the evaluated properties. Among these, base model knowledge ($\Delta_{\text{base}}$) exhibits the strongest correlation.}
    \label{fig:experiments_figure}
\end{figure*}

To address this question, we define the notion of \emph{mergeability}: a property of model updates that captures how well they retain trained knowledge when merged with other model updates. We find that not all model updates have the same mergeability---there is a wide spectrum of mergeability among models. We use the term mergeability score to quantify this property.

This work investigates possible sources for having high or low mergeability. 
Figure \ref{fig:experiments_figure} describes our experimental setup. Upon obtaining model updates (we use LoRA adapters \cite{hu2022lora} in our experiments), we calculate each model update's mergeability score (the degree to which the knowledge encoded in a given model update is preserved when it is merged with other model updates. For a given model update, we merge it with a subset of randomly sampled other model updates and evaluate the merged model on the given model update task. We repeat this multiple times and measure the average performance; further described in \S\ref{sec:mergeability}). We then group model updates by their mergeability score and investigate different causes for high mergeability: 
 general domain knowledge of the base model, weight properties, and specific task knowledge of the base model.

Our study spans two experimental setups: example-level mergeability using the PopQA dataset \cite{mallen-etal-2023-trust} and task-level mergeability using the Lots-of-LoRAs collection \cite{bruel2024compress}. In the PopQA dataset, we find that a low probability gap between the top predicted answer and the correct answer in the base model correlates with a higher mergeability. This suggests that the base model’s prior knowledge about a question strongly influences whether fine-tuning on knowledge related to this question will merge better. Consistently, in Lots-of-LoRAs, we observe that tasks where the base model initially performed well suffer less performance degradation from merging, whereas tasks on which the base model had lower initial accuracy suffer from larger drops in performance after merging.
Our findings indicate that the mergeability of fine-tuned models is closely linked to the base model’s initial performance on the corresponding fine-tuning data. Other possible causes, namely, general domain knowledge or structural weight properties, do not correlate well with mergeability scores. 

In further analyses, we discover that mergeability is primarily a local trait of the model update and does not depend on the merged set. We show that when merging a highly mergeable model update with model updates from other mergeability groups, the highly mergeable model update remains stable regardless of the partner group we merge with.

Finally, as a proof-of-concept application of our insights, we propose a simple merging technique that incorporates the base model's performance in a weighted mean merging. We show that this technique improves the retention of weak-performing tasks with little or no degradation of strong-performing tasks. 

Our contributions in this paper are threefold: 
\begin{itemize}
    \item We show the existence of mergeability and propose a concrete, measurable definition of it.

    \item We provide empirical evidence that the base model’s prior knowledge is a key predictor of the mergeability of fine-tuned weights. To our knowledge, this is the first study to directly link pre-training knowledge with mergeability.

    \item We show an application of our findings and suggest merging weights with awareness of the base model performance.
\end{itemize}
\section{Mergeability}
\vspace{-4pt}
\label{sec:mergeability}

\emph{Mergeability} is a trait of model updates describing the degree to which the knowledge encoded in a model update is preserved when merged with other model updates. This falls into a larger context of what allows successful merging (e.g., high-dimensionality and a shared base model;  \S\ref{sec:related}), but is unique in recognizing the effect of the model updates themselves. In this section, we define and empirically show the existence of mergeability.

\paragraph{Mergeability score.}
Given a model update $\theta_{\Delta}$ with corresponding input-output pair $(x,y)$ and a distribution $\mathcal{D}$ over sets of other updates,
we define the mergeability score $S$ of $\theta_{\Delta}$ as:
\begin{equation}
\resizebox{0.99\linewidth}{!}{$
    S(\theta_{\Delta}) = \mathbb{E}_{\{\theta_{\Delta j}\}\sim\mathcal{D}} 
    \big[ f\big(\theta + \mathcal{M}(\{\theta_{\Delta}\}\cup \{\theta_{\Delta j}\}) ; x, y \big) \big] 
    $}
\end{equation}
where $f(\theta; x, y)$ is a scoring function that evaluates a model with parameters $\theta$ on a set of input-output data pairs $(x,y)$ and $\mathcal{M}$ is a merging algorithm that combines a set of updates into a single update. 

To estimate the mergeability score $S(\theta_{\Delta})$, we conduct $N$ trials.
In each trial $i{\in}\{1,\dots,N\}$, we sample $M$ other model updates
$\{\theta_{\Delta m}\}_{m=1}^M$ (without replacement) from the global pool
and merge $\{\theta_{\Delta}\}\cup \{\theta_{\Delta m}\}_{m=1}^M$,
yielding a merged model update $\theta_\Delta^{(i)}=\mathcal{M}(\{\theta_{\Delta}\}\cup \{\theta_{\Delta m}\}_{m=1}^M)$.
We then update the base model with $\theta_\Delta^{(i)}$ and re-evaluate,
obtaining the following empirical mergeability score:
\begin{equation}
S(\theta_{\Delta e}) = \frac{1}{N} \sum_{i=1}^{N} f(\theta+\theta_\Delta^{(i)}; x,y)
\end{equation}

This score captures the robustness of an example’s knowledge under merging, where higher values indicate stronger mergeability with other model updates. The scoring function $f$ depends on the setup, as detailed in the next section. 

\begin{figure}[t]
    \centering
     \includegraphics[width=1\linewidth]{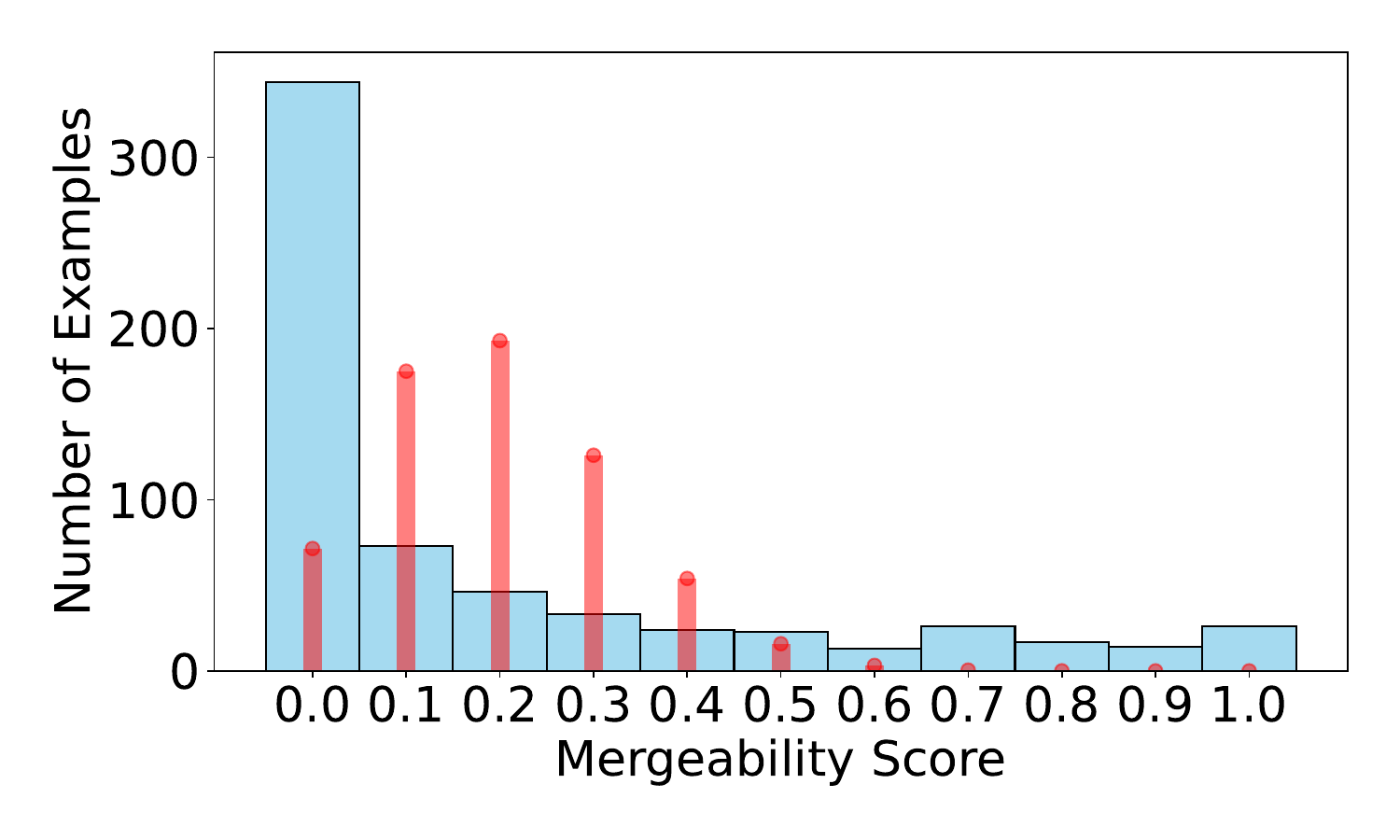}
    \caption{Mergeability score distribution of Llama-3.2-3B on the PopQA dataset. Blue wide bars show the mergeability score as empirically calculated. Red thin bars show the baseline distribution if mergability were not a model trait, modeled as a binomial distribution with a fixed success rate.}
    \label{fig:llama_mergeability_10}
\end{figure}

In practice, not all model updates are equally stable under merging: some integrate seamlessly into a merged model, while others tend to interfere or degrade. Figure \ref{fig:llama_mergeability_10} shows the mergeability distribution of Llama \cite{dubey2024llama} on the PopQA dataset (more distributions are in Appendix \ref{sec:additional_distributions}). 
The blue bars show the mergeability score as empirically calculated in our experiments. The red bars show the distribution as if the scores were distributed randomly as a binomial distribution with a success probability of $P = \frac{\text{\# of success merges}}{\text{Total \# of merges}}$, where we consider a merge as succeeded if it retained the model update information after merging. The difference in the distributions shows the existence of non-trivial mergeability.

\paragraph{Research questions.}
After establishing that mergability is a non-trivial property, we investigate several possible reasons for mergeability:
\begin{itemize}
    \item Base model specific task knowledge (\S\ref{sec:rq1}).

    \item Weight properties (\S\ref{sec:rq2}).    
    \item Base model general domain knowledge (\S\ref{sec:rq3}).
\end{itemize}

Then, in \S\ref{sec:other} we check if mergeability is a local trait of the model update, or rather depends on the group of model updates it is merged with.

\vspace{-5pt}
\section{Experimental Setup}
\vspace{-5pt}
\label{sec:setup}
We evaluate the mergeability of LoRA adapters \cite{hu2022lora} as recent work showed that such merging is very powerful \cite{stoica2024model}. We focus on two settings: (i) entity-centric question answering with PopQA \cite{mallen-etal-2023-trust} dataset, and (ii) a broad collection of LoRA adapters from \textit{Lots-of-LoRAs} collection \cite{bruel2024compress}. 
The first corresponds to example-level mergeability, merging adapters capturing single data points, while the second corresponds to task-level mergeability. 
Unless otherwise specified, all merges are performed with \textsc{Knots} \cite{stoica2024model} merging algorithm, which is specifically designed for merging LoRA adapters.\footnote{We report analyses results with other merging algorithms in Appendix \ref{sec:additional_merging_algo}, finding mostly consistent patterns.} 
In each of the two setups, we describe our instantiations of the mergeability score and the evaluation protocol;\footnote{Additional experiments hyperparameter details are available at Appendix \ref{sec:additional_parameters}}

\vspace{-1pt}
\subsection{Example-Level Mergeability: PopQA}
\vspace{-1pt}
PopQA is an open-domain, entity-centric QA benchmark \cite{mallen-etal-2023-trust}, making it suitable for studying example-level mergeability. 
In this case, the scoring function $f$ is chosen as the binary correctness of the model:
\begin{equation}
f(\theta; x,y) = \mathbbm{1}\{\hat{y} = y\},
\end{equation}
where $\hat{y}$ is the model prediction.

For controlled probabilistic evaluation, we convert it to a multiple-choice format: for each question with gold answer $y$, we sample $n$ additional candidates from answers to other PopQA questions (without replacement). To reduce spurious ambiguity, we discard distractors that are string-identical to $y$ or that are near-duplicates by normalization (case/punctuation stripping). We use $n{=}7$ (and obtain 8-option multiple-choice questions).


 \paragraph{Models.}
We use two large pretrained language models from distinct families: Llama-3.2-3B \cite{dubey2024llama} and Qwen-2.5-3B \cite{hui2024qwen2}. This choice allows us to compare models from different families, providing a more robust testbed for mergeability. Results in the main paper are reported for the Llama-3.2-3B model; additional analyses, including results on Qwen-2.5-3B, are provided in Appendix~\ref{sec:additional}.

 \paragraph{Evaluation.} 
 We evaluate with $k$-shot prompting \cite{brown2020language}. In our experiments, we used $k=4$. Given the prompt and a set of options $\{y_j\}_{j=1}^{n+1}$, with the correct answer $y$ among them, we compute the length-normalized conditional log-probability assigned by the model to each option:
 \begin{equation}
     \text{score}(y_j) \,=\, \frac{1}{|y_j|} \sum_{t=1}^{|y_j|} \log p_\theta\!\left((y_j)_t \,\middle|\, \text{prompt}, (y_j)_{<t}\right)
 \end{equation}

We predict $\hat{y}=\arg\max_j \text{score}(y_j)$. Normalization mitigates length bias across options.

\paragraph{Training Procedure.}  
To focus on model updates that change the model’s knowledge, we first filter out questions already answered correctly by the base model under the above evaluation protocol. For the remaining (incorrect) questions, we retrieve the relevant entity’s Wikipedia page. We then fine-tune the model with LoRA on these entity-related passages to teach the base model relevant information on the questioned entity. We retain only those examples for which the post-finetuning model answers correctly. This yields a collection of example-specific LoRA adapters that each corrects a distinct factual error of the base model. Out of $1931$ examples we tested, $1107$ were answered correctly without training, which left us with $824$ examples we trained ($42.67\%$). Out of them, for $639$ ($77.55\%$) examples the correct answer had the highest probability after training. Additional finetuning details are provided in Appendix~\ref{sec:additional_parameters}. We also report experimental results with additional LoRA ranks (Appendix \ref{sec:lora_rank_affect}) and full finetuning (Appendix \ref{sec:full_finetuning}). Trends are consistent with the main paper results.

\subsection{Task-Level Mergeability: Lots-of-LoRAs}
To study task-level mergeability, we use the Lots-of-LoRAs benchmark \cite{bruel2024compress}, a large-scale collection of LoRA adapters trained on diverse NLP tasks.
In this case, the scoring function $f$ is computed as the post-merging task accuracy:

\begin{equation}
    f(\theta; x, y) \,=\;\text{Acc}^{(i)}_{\text{merged}}(x, y)
\end{equation}


\begin{figure*}
    \centering
    \includegraphics[width=0.75\linewidth]{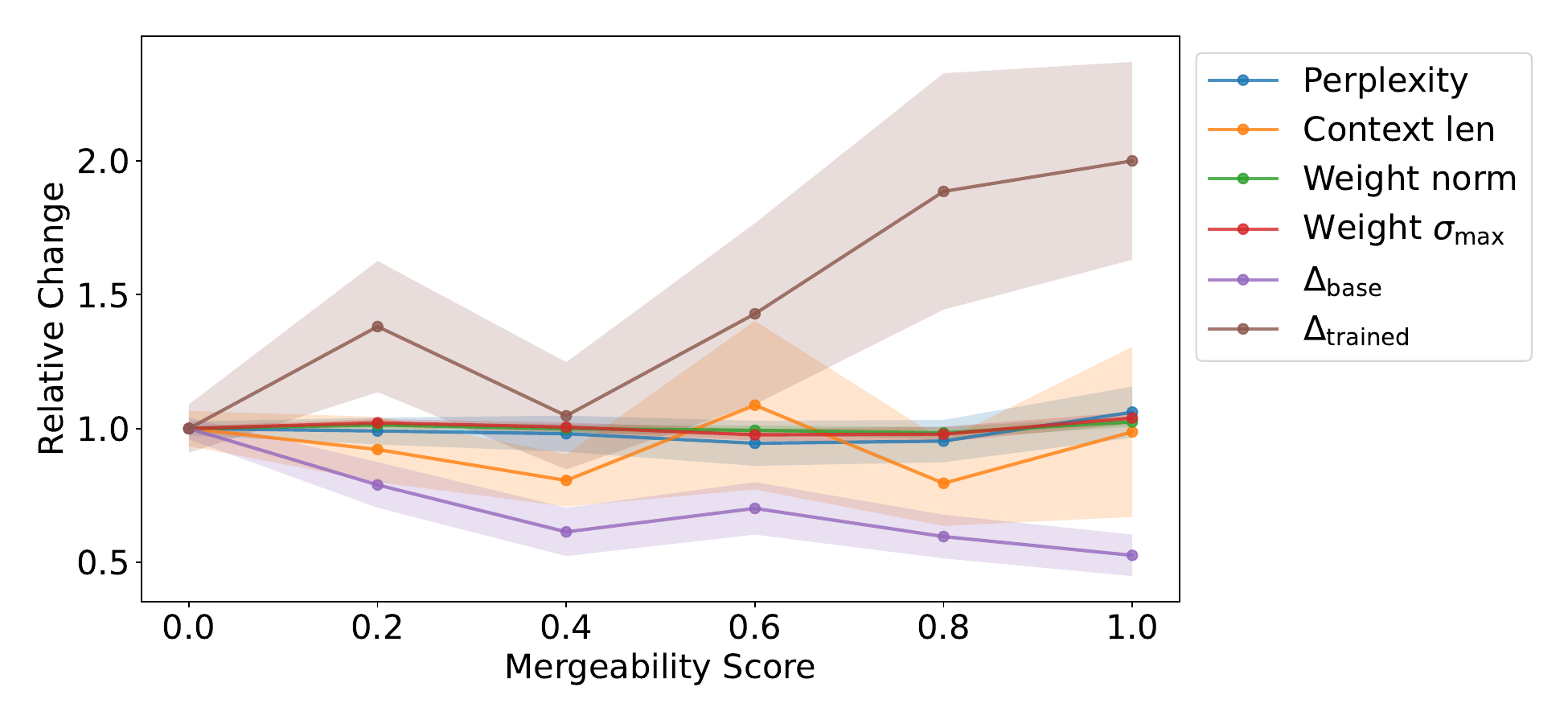}
    \caption{PopQA different properties correlation with mergeability score results. Values are mean values normalized relative to $S=0.0$. $\Delta_{\text{base}}$ trend shows that the gap decreases with mergeability, implying that examples with better base model knowledge are more mergeable. Additionally, $\Delta_{\text{trained}}$ shows that high-mergeability examples achieve larger post-training probability improvements, although they were easier to ``fix''. Conversely, we observe no clear trend between mergeability score and training data difficulty (average perplexity and average context length) or weight-level properties (average weight norm and average highest singular value). The shaded regions represent the standard error of the values.}
    \label{fig:popqa_all_results}
\end{figure*}

 \paragraph{Model.}  
We use the same base model as in \citet{bruel2024compress}, Mistral-7B-Instruct-v0.2 \cite{jiang2023mistral7b}.

\paragraph{Evaluation.}  
For each task, we use the test prompts provided by the dataset. We evaluate by exact match (EM) at the example level and aggregate to task accuracy. We evaluate on each task's test set. To control for ceiling effects, we restrict evaluation to tasks where the finetuned model achieves at least $99\%$ accuracy, ensuring that mergeability is measured between adapters that individually solve their target tasks near perfectly. This left us with a total of $81$ tasks (We provide experimental results for other accuracy thresholds in Appendix \ref{sec:lots_of_loras_additional}, trends are consistent across all thresholds).

\vspace{-5pt}
\section{Causes of Mergeability}
\vspace{-5pt}
\label{sec:mergeability_causes}

In this section, we investigate possible causes for high or low mergeability. In each case, we bin models by their mergeability score and correlate these scores with metrics reflecting possible causes. 

\subsection{What is the effect of the base model's knowledge of the task?}
\label{sec:rq1}
In this research question, we wish to study how the base model’s prior knowledge of an example/task impacts mergeability.



In example-level PopQA analysis, Figure~\ref{fig:llama_answer_rank} reports the average rank of the correct answer under the base model. A higher rank (closer to zero) indicates that the base model assigns the correct answer a better rank. While we generally observe that higher-mergeability examples correspond to lower ranks, we also observe a non-monotonic jump for $S=1.0$. 

We further analyze the probability gap between the most likely answer and the correct answer in the base model $\Delta_{\text{base}} = p_{\max}^{\text{base}} - p_{\text{correct}}^{\text{base}}$ (Figure~\ref{fig:popqa_all_results}).
The gap decreases with mergeability, implying that examples requiring only a small adjustment to the decision boundary are more mergeable. Moreover, when we compare base vs.\ post-training gaps $\Delta_{\text{trained}}=p_{\text{correct}}^{\text{trained}}-p_{\text{correct}}^{\text{base}}$ (Figure~\ref{fig:popqa_all_results}), we find the opposite trend: high-mergeability examples achieve larger post-training probability improvements. This indicates that examples that were “easier” to fix also produce more stable model updates under merging.

\begin{figure}[t]
    \centering
    \includegraphics[width=0.9\linewidth]{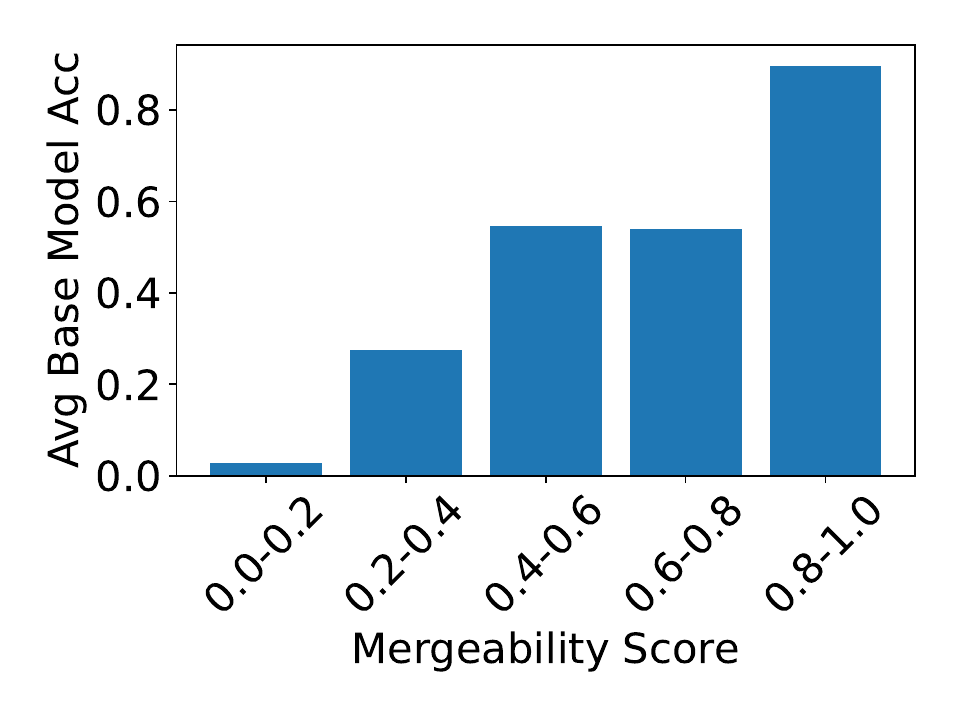}
    \caption{Lots-of-LoRAs average base model task accuracy of different mergeability scores. Higher mergeability scores have on average a higher base model accuracy.}
    \label{fig:lots_of_lora_merge_vs_base_acc}
\end{figure}

Lots-of-LoRAs analysis reveals similar trends on the task-level merging. As Figure \ref{fig:lots_of_lora_merge_vs_base_acc} shows, higher mergeability scores have higher average base model accuracy. In other words, tasks with better base model knowledge of the task have better mergeability scores. 

Overall, we conclude that \textbf{better knowledge of the task or fine-tuning data in the base model corresponds to higher mergeability}. 
 
\subsection{Does training data difficulty affect mergeability?}
\label{sec:rq2}

To examine whether mergeability is also affected by training data difficulty, we consider two proxies: base model perplexity and length (token count) of the training context.

As Figure~\ref{fig:popqa_all_results} shows, there are no clear trends w.r.t mergeability score in the case of example-level mergeability in PopQA. 
One exception is a high base model perplexity on examples with the highest mergeability score ($S=1.0$). This suggests that model updates correcting knowledge gaps in regions where the base model is highly uncertain are more robust to merging. However, overall perplexity and context length are poor predictors of mergeability in this case. 

In contrast, the Lots-of-LoRAs analysis reveals different trends. As Figure \ref{fig:lots_of_lora_perplexity_vs_mergeability} shows, higher-mergeability examples correspond to lower perplexities. Moreover, Figure \ref{fig:lots_of_lora_train_lenght_vs_mergeability} shows that longer training contexts have higher mergeability scores. 

However, a difference in the two settings must be noted. While PopQA training data are general passages on an entity, with a different format from the evaluated questions, Lots-of-LoRAs training data is highly related to the evaluation data, sharing the same task format. This suggests that we can observe training data effects on mergeability only when it is aligned with the evaluation data.

\subsection{Do weight-level properties correlate with mergeability?}
\label{sec:rq3}
We next examine whether structural properties of the learned weight updates correlate with mergeability.
We compute two structural metrics on the effective update matrix $\Delta W = BA$ (where $A, B$ are LoRA update matrices): (i) Frobenius norm $\|\Delta W\|_F$, previously shown to influence mergeability \cite{pari2025collective, horoi2025less} and (ii) highest singular value $\sigma_{\max}(\Delta W)$, which captures dominant update directions \cite{stoica2024model}.

\begin{table}[]
\centering
\begin{tabular}{lcc}
\hline
\multicolumn{1}{c}{\begin{tabular}[c]{@{}c@{}}Mergeability\\ Score\end{tabular}} &
  \begin{tabular}[c]{@{}c@{}}Avg Weight \\ Norm\end{tabular} &
  \begin{tabular}[c]{@{}c@{}}Avg Weight\\  $\sigma_{\max}$\end{tabular} \\ \hline
0.0-0.2 & 1.15 & 0.78 \\
0.2-0.4 & 0.57 & 0.40 \\
0.4-0.6 & 0.73 & 0.52 \\
0.6-0.8 & 0.65 & 0.46 \\
0.8-1.0 & 0.66 & 0.46 \\ \hline
\end{tabular}
\caption{Lots-of-LoRAs results for weight properties. We do not observe a clear correlation between weight properties and the mergeability score. However, we observe a notable distinction between extremely low mergeability examples ($S\in[0.0, 0.2)$) and all other scores ($S\geq0.2$). Examples in the lowest bin exhibit significantly higher average norms and singular values.}
\label{tab:lots_of_lora_weight_properties}
\end{table}

In the PopQA setting (Figure~\ref{fig:popqa_all_results}), we observe that the highest mergeability examples tend to have slightly higher norms and higher $\sigma_{\max}$. 
However, overall the correlations to mergeability are close to zero ($0.10$ and $0.09$ Spearman's). 
The Lots-of-LoRAs analysis (Table~\ref{tab:lots_of_lora_weight_properties}) again shows no consistent trend across mergeability scores. However, we observe a notable distinction between extremely low mergeability examples ($S\in[0.0, 0.2)$) and all other scores ($S\geq0.2$). Examples in the lowest bin exhibit significantly higher average norms and singular values. 

The difference in trends between PopQA and Lots-of-LoRAs might be explained by the difference in training setting - single model parameter training (mlp up proj) and single layer training in PopQA versus multiple model parameters (attention Q, K, and V matrices) and multiple layer training in Lots-of-LoRAs. However, in both experiments \textbf{we do not observe a clear correlation between weight properties and mergeability score}.

\vspace{-7pt}
\section{Other Mergeability Properties}
\vspace{-10pt}
\label{sec:other}

Having examined potential causes of mergeability, we next turn to complementary aspects of this phenomenon: properties that characterize how mergeability behaves under different conditions.

\subsection{Is mergeability local or global?}


\begin{figure}
    \centering
    \includegraphics[width=1\linewidth]{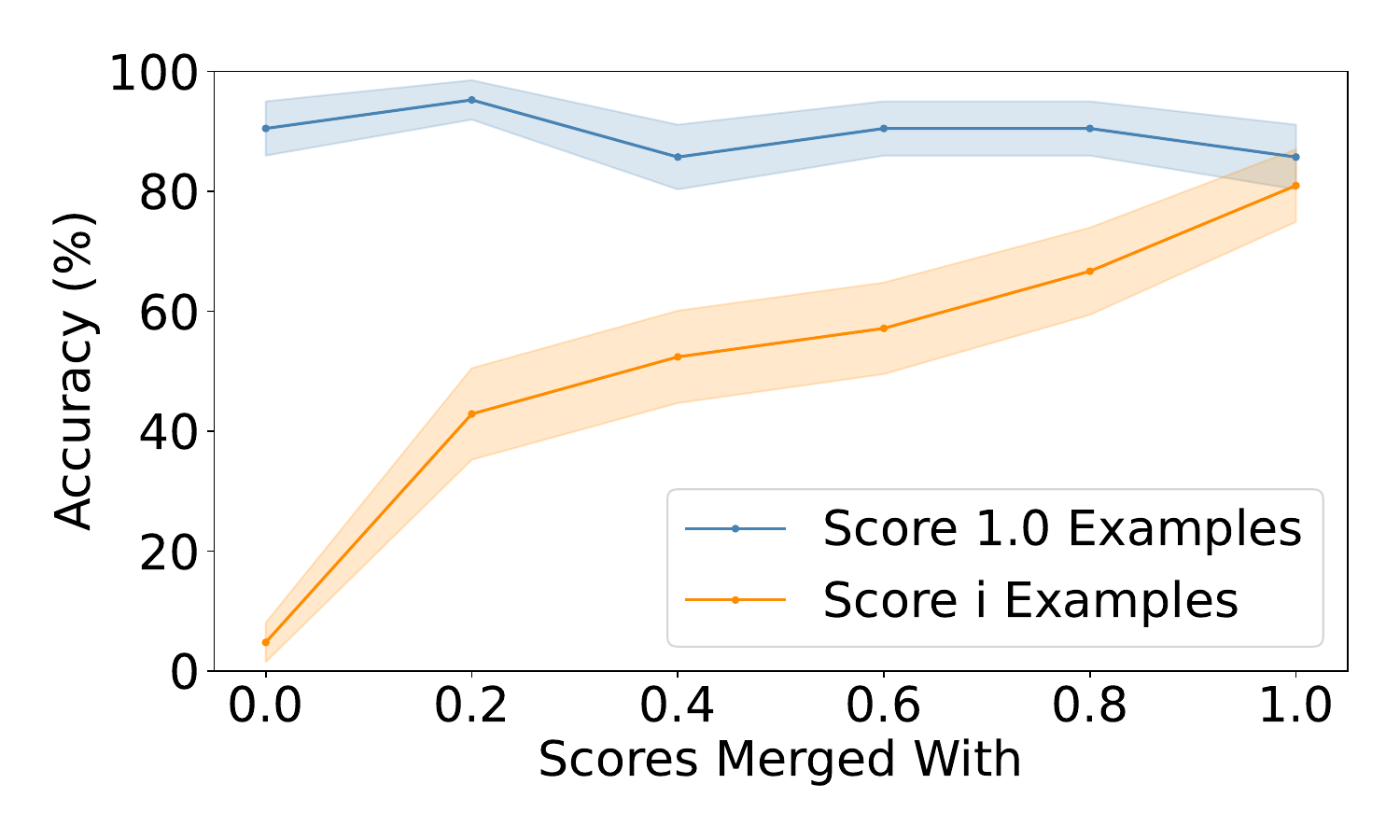}
    \caption{The mergeability score when merging weights $\theta_\Delta$ with mergeability score $S(\theta_\Delta)=1.0$ with weights drawn from different mergeability scores. The blue line, which represents examples with a fixed score $S(\theta_\Delta)=1.0$, shows near-constant accuracy across conditions, while the orange line (other mergeability group examples) improves with their own mergeability score. The shaded regions represent the standard error of the accuracy.}
    \label{fig:local_mergeability}
\end{figure}

Previous work has shown that merging is affected by the merge set, and more specifically, by the shared knowledge between tasks \cite{zaman2023fuse}. However, what happens when no shared knowledge exists?
To test whether mergeability is an intrinsic property of a model update or depends on the merge set, we conduct a controlled experiment under the PopQA setting. We fix a set of highly mergeable model updates ($S(\theta_\Delta)=1.0$) and merge them with groups drawn from bins of varying mergeability scores. If mergeability is a local trait depending only on the target model update and not on the merge set, then performance on the fixed $S(\theta_\Delta)=1.0$ updates should remain stable regardless of the partner group. Figure~\ref{fig:local_mergeability} confirms this hypothesis: blue bars (fixed $S(\theta_\Delta)=1.0$ updates) show near-constant accuracy across conditions, while orange bars (other mergeability group updates) improve with their own mergeability score.

This finding suggests that mergeability is primarily an intrinsic property of the LoRA update itself rather than an emergent property of the merge set.

\subsection{How merging algorithm affects mergeability?}
\label{sec:merging_affect}
Our experiments primarily employed \textsc{Knots} \cite{stoica2024model}, the current state-of-the-art algorithm for merging LoRA weights. To assess how the choice of merging algorithm influences mergeability, we compare \textsc{Knots} with two other merging algorithms: TIES \cite{yadav2023ties} and simple mean averaging.

\begin{figure}[t]
    \centering
    \includegraphics[width=1\linewidth]{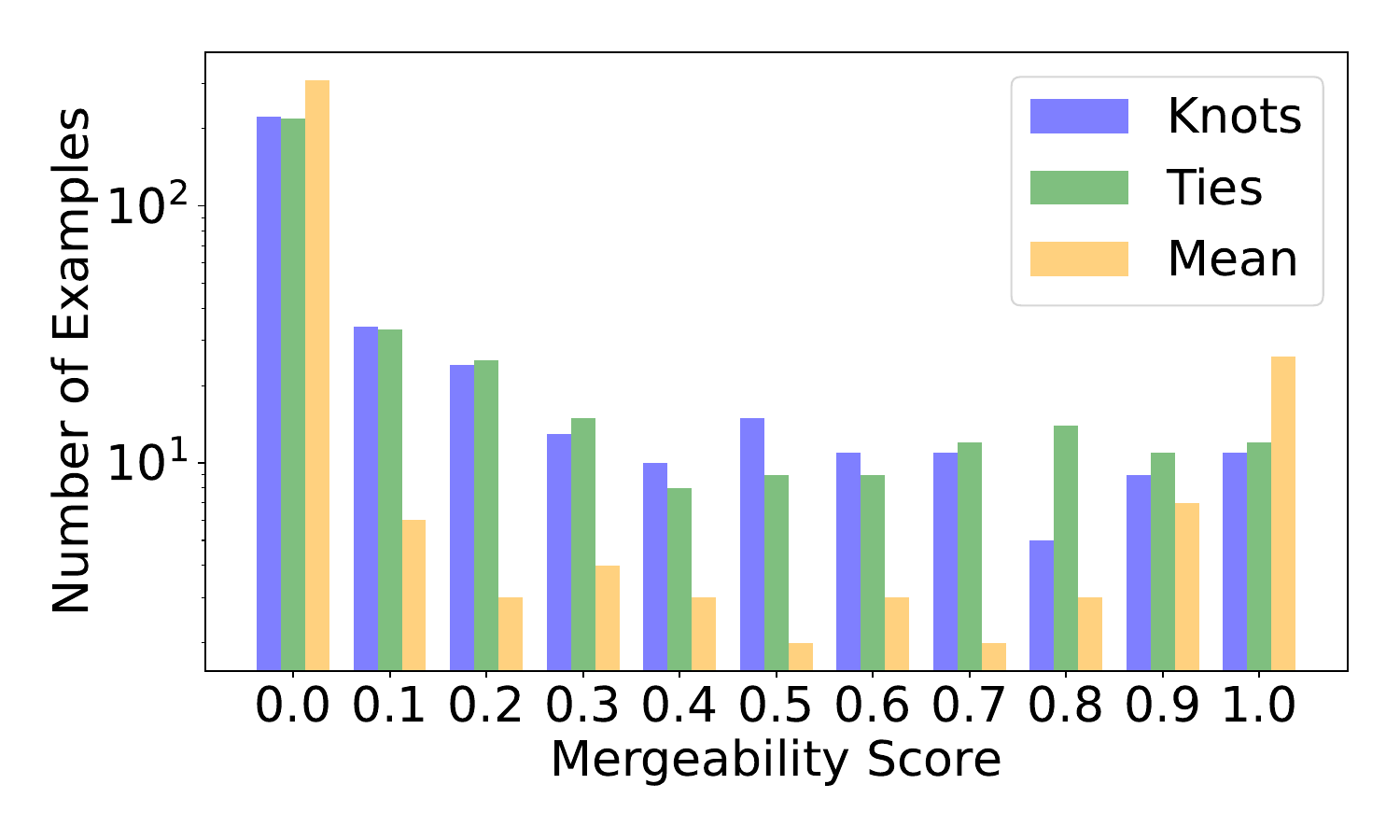}
    \caption{Mergeability score distribution for the Qwen model on PopQA using different merging algorithms (log scale). As the method is `weaker', more examples are in the higher bins.}
    \label{fig:different_merging_algorithms}
\end{figure}

Figure~\ref{fig:different_merging_algorithms} presents the mergeability score distribution for the Qwen model on PopQA using these merging algorithms. A clear trend emerges: ``weaker'' algorithms exhibit more examples in the higher mergeability bins. Specifically, TIES yields more examples with scores $S(\theta_\Delta)\geq 0.8$ than \textsc{Knots}, while mean averaging produces the largest number of examples at $S(\theta_\Delta)=1.0$. 

This pattern reflects the degree of interference resolution. Mean averaging, the weakest method, performs no conflict mitigation—weights tend to either merge or fail. TIES introduces global sign conflict resolution, and \textsc{Knots} further aligns update subspaces, improving overall merging. However, these interventions slightly reduce the proportion of perfectly mergeable examples, suggesting a trade-off between resolving interference and preserving higher mergeability.

\vspace{-5pt}
\section{Mergeability Score for Better Merging}
\vspace{-5pt}
Our analysis of mergeability revealed a consistent relationship between a base model's task accuracy and the corresponding model update mergeability (\S\ref{sec:mergeability_causes}). Specifically, model updates corresponding to tasks where the base model already performs well tend to have a better mergeability than model updates of tasks with lower base model accuracy. This suggests that naive averaging of adapter may  overemphasize high-accuracy tasks, while underweighting tasks that have lower mergeability.

To address this imbalance, we propose a weighted averaging strategy where the contribution of each model update is inversely proportional to the base model’s accuracy on the corresponding task. The intuition is to assign a higher weight to model updates that have lower mergeability, while limiting the influence of model updates on tasks for which the base model already works well. 

Let $\{\Theta_{\Delta1}, \Theta_{\Delta2}, \dots, \Theta_{\Delta T}\}$ be a set of model updates corresponding to tasks $\{t_1, t_2, \dots, t_T\}$. For each task $t_i$, we compute the base model accuracy $Acc(t_i) \in [0, 1]$. We then define the inverse accuracy score as $s_i = 1 - Acc(t_i)$. 
To convert these scores into weights, we apply a softmax function with temperature $\tau$:
$
w_i = \frac{\exp(s_i / \tau)}{\sum_{j=1}^{T} \exp(s_j / \tau)}
$.
The final merged adapter is computed as a weighted sum:
\begin{equation}
    L_{\text{merged}} = \sum_{i=1}^{T} w_i \cdot \Theta_{\Delta i}
\end{equation}

This approach ensures that tasks where the base model performs poorly (i.e., high $s_i$) receive higher weights, while tasks with a high base accuracy are down-weighted. The temperature parameter $\tau$ controls the sharpness of the weighting distribution: lower values of $\tau$ emphasize differences more strongly.

\begin{figure}
    \centering
    \includegraphics[width=0.9\linewidth]{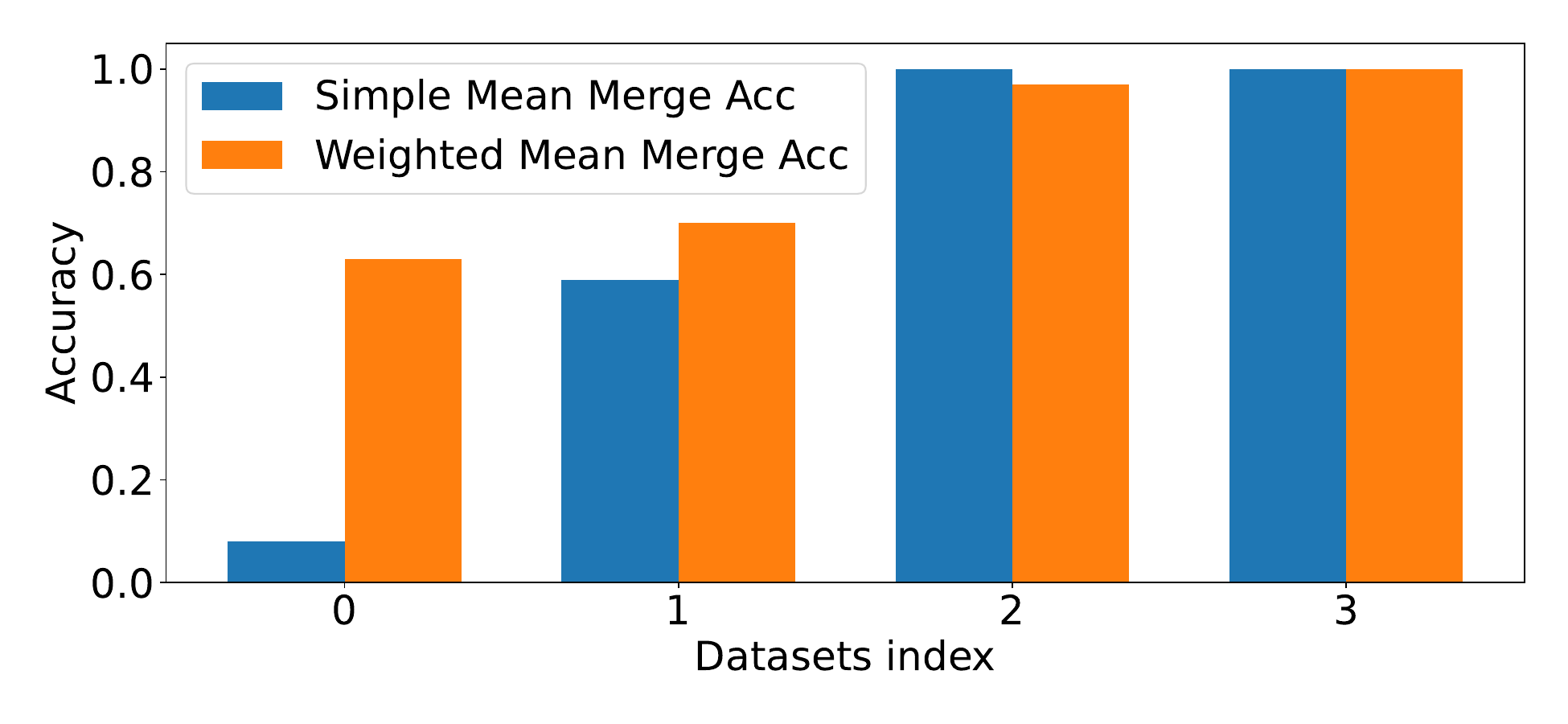}
    \caption{Simple mean merging versus weighted mean merging based on base model accuracy. Weighted average merging (orange bars), low-accuracy tasks (tasks 0 and 1) retained more of their fine-tuned performance, while high-accuracy tasks experienced minimal degradation (task 3) or no degradation at all (task 4).}
    \label{fig:method_results}
\end{figure}

To evaluate our method, we used the Lots-of-LoRAs collection. We sampled 2 random tasks with a low base model accuracy and 2 tasks with a high base model accuracy, and merged their model updates using a weighted average as described above. As shown in Figure \ref{fig:method_results},  in the weighted average experiment (orange bars), low-accuracy tasks (tasks 0 and 1) retained more of their fine-tuned performance, while high accuracy tasks
experienced minimal degradation (task 3) or no degradation at all (task 4). In regular averaging (blue bars), tasks with lower base model accuracy experienced a higher performance degradation. 
\section{Related Work}
\vspace{-4pt}
\label{sec:related}
\paragraph{Model merging.}
Model merging has emerged as a technique for combining fine-tuned models (each expert on a different task) into a single, multitask model without requiring additional training. Direct approach is a simple average of the weights. \citet{wortsman2022model} proposed Model Soups, finding that averaging the weights of multiple fine-tuned variants can improve accuracy and out-of-distribution robustness over the individual models. 

Another line of work addresses direct conflicts between model parameters. \citet{yadav2023ties} propose TIES-Merging, which identifies and resolves conflicting parameter updates (differences in sign or scale for the same weight in different models) prior to merging. By correcting such sign mismatches and scaling disparities, their method reduces destructive interference, leading to merged models that suffer smaller drops in accuracy. In the context of parameter-efficient fine-tuning, \citet{stoica2024model} focuses on merging LoRA adapter weights. They found that averaging LoRA deltas is challenging due to misaligned update subspaces, and introduced an SVD-based alignment (\textsc{Knots}) to transform each model’s adapter weights into a common basis before merging. This alignment significantly improved the compatibility of fine-tuned LoRA weights.
\citet{ilharco2022editing} explores model merging using task vectors. They define a task vector as the difference between a fine-tuned model’s weights and the original base model’s weights. These task vectors can then be added, subtracted, or scaled.

\paragraph{What helps mergeability?}
Prior work has begun to explore this phenomenon. \citet{zaman2023fuse} showed that unshared knowledge between tasks is retained less during merging. They observe that when models are merged, information that was learned by all the experts is usually preserved in the fused model, whereas information unique to a single model is prone to being overwritten or forgotten.
The importance of aligned representations and update directions has also been noted. 
Some works looked at the effect of the chosen base model on mergeability. \citet{yadav2024matters} and \citet{he2025mergebench} show that it is easier to merge bigger and stronger base models. Other works considered how the process of training itself affects merging, suggesting that more training, which shows larger norm updates \citep{gueta2023knowledge} leads to worse merging performance \citep{pari2025collective,horoi2025less}. 
\citet{stoica2024model} highlighted that merging LoRA fine-tunings can fail when the two models’ updates reside in different singular directions, effectively lacking a shared basis. \citet{ainsworth2022git,jordan2023repair} similarly point out that two networks might need a permutation alignment to correspond to the same functions. These works suggest that when models make fundamentally different internal choices, naive merging will cause interference.

While all mentioned works at least implicitly point to what makes a good merging outcome, we are the first to notice specific models are merged better even under similar training conditions and as an outcome, the first to demonstrate reasons for that phenomenon.

\paragraph{Affinity outside model merging.} In fields such as multitask learning \citep{bingel-sogaard-2017-identifying,kim-etal-2023-taskweb}, continual and intermediate training \citep{poth-etal-2021-pre}, many works have  studied which tasks aid each other or, in general, which curriculum is beneficial \citep{hacohen2019power, shrivastava2016training}. There, a common theme is discussing how a specific task might help another (e.g., a low-resource language improved by training on English \citealp{Bansal2018PretrainingOHA}). Fewer works point at universality, as we do, where specific models are inherently good. Those include works on fine-tuning \citep{choshen-etal-2023-start} and reinforcement learning \citep{guo2025deepseek}, but perhaps most known is pretraining itself, which is shown to be a strong start for many counterparts \citep{devlin2019bert,aryabumicode}. 

\vspace{-5pt}
\section{Conclusions}
\vspace{-5pt}
In this work, we show the existence of and analyze \emph{mergeability} - a property of the model updates that quantifies the robustness of model updates when merged with other model updates. Our analysis across both example-level and task-level settings, demonstrates that mergeability is not uniformly distributed: some model updates have a better mergeability than others. We raised different key possible causes for mergeability: base model general domain knowledge, weight properties, and base model specific task knowledge. To investigate this property, we define a \emph{mergeability score}, which allows us to measure mergeability and study what affects it. We find that base model task knowledge is correlative with mergeability -- instances with higher knowledge in the base model are more mergeable. Moreover, we find evidence that mergeability is an intrinsic property of the model update itself, and not a property of the merging set. Finally, we illustrated how mergeability scores can guide improved merging strategies, mitigating imbalances between tasks with differing base model familiarity.

\section*{Limitations}
While this work is based on an extensive set of experiments, several limitations are worth noting and can be addressed in future research.
First, our approach can be readily extended to evaluate additional datasets, tasks, and base models to further verify the consistency of our core findings.
Second, although our analysis focuses on merging LoRA adapters, it can naturally be extended to other types of tuned models, from full-model fine-tuning to alternative adapter architectures and even local model updates.
Third, while we use \textsc{Knots} as our primary merging algorithm, other approaches such as Ties and Mean merit further exploration.

\section*{Ethical considerations}
Our work adds to the body of literature on model merging and might help develop better merging algorithms. We do not foresee major risks associated with this work. However, a malicious actor
might use our analysis to better understand how to amplify unwanted behaviours during model merging.

\section*{Acknowledgments}
This research was supported by an Azrieli Foundation
Early Career Faculty Fellowship, Open Philanthropy, and by an IBM-Technion Research Collaboration.  This research was funded by
the European Union (ERC, Control-LM, 101165402). Views and opinions expressed are however
those of the author(s) only and do not necessarily reflect those of the European Union or the European
Research Council Executive Agency. Neither the European Union nor the granting authority can be
held responsible for them.

\bibliography{custom}

\newpage
\appendix
\section{Appendix}

\subsection{Additional Results for \S\ref{sec:mergeability_causes}}
\label{sec:additional}





\begin{figure}[h]
    \centering
    \includegraphics[width=1\linewidth]{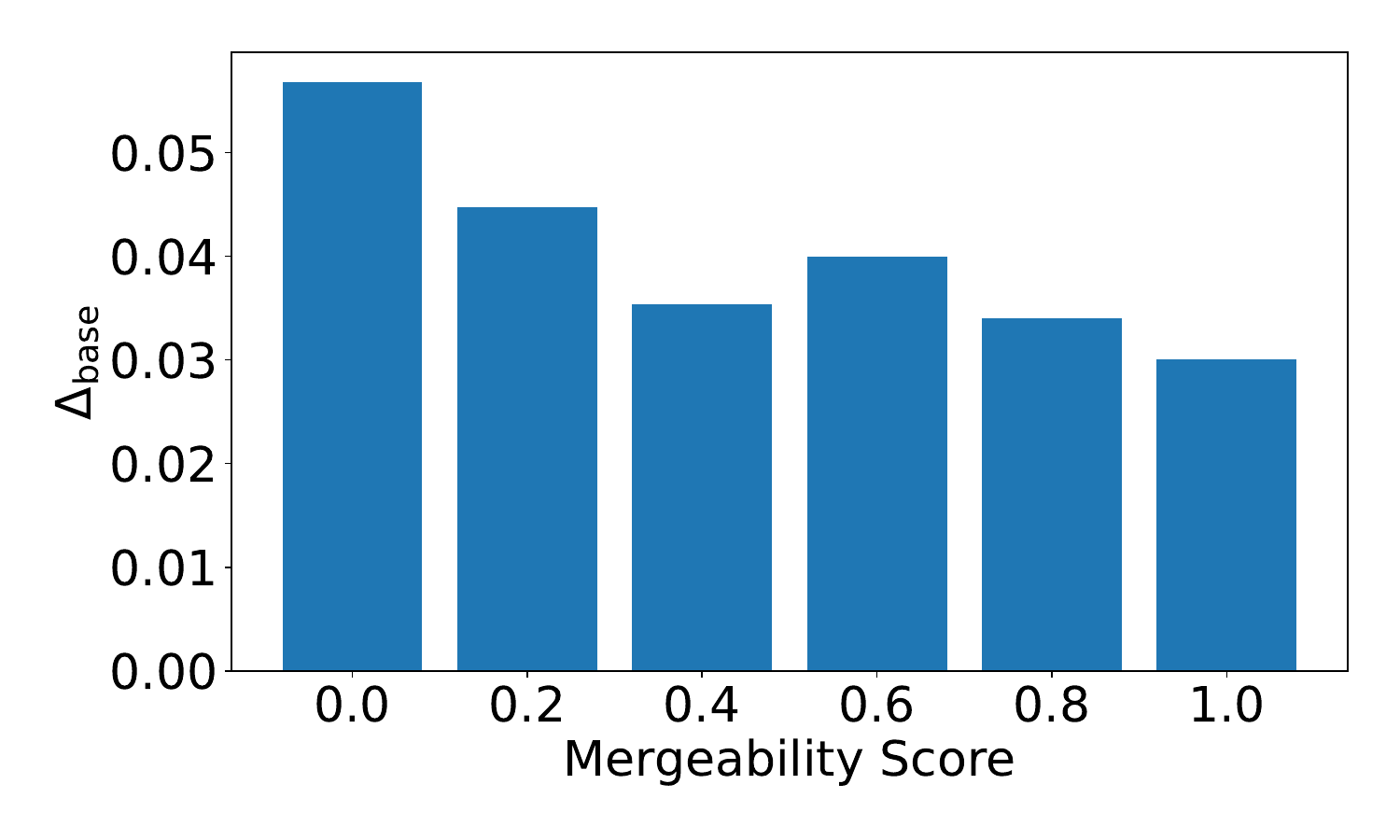}
    \caption{PopQA average difference between the highest and the correct answer probability in the base model. We observe that the gap decreases with mergeability, implying that examples with better base model knowledge are more mergeable.}
    \label{fig:llama_highest_correct_diff}
\end{figure}

\begin{figure}
    \centering
    \includegraphics[width=1\linewidth]{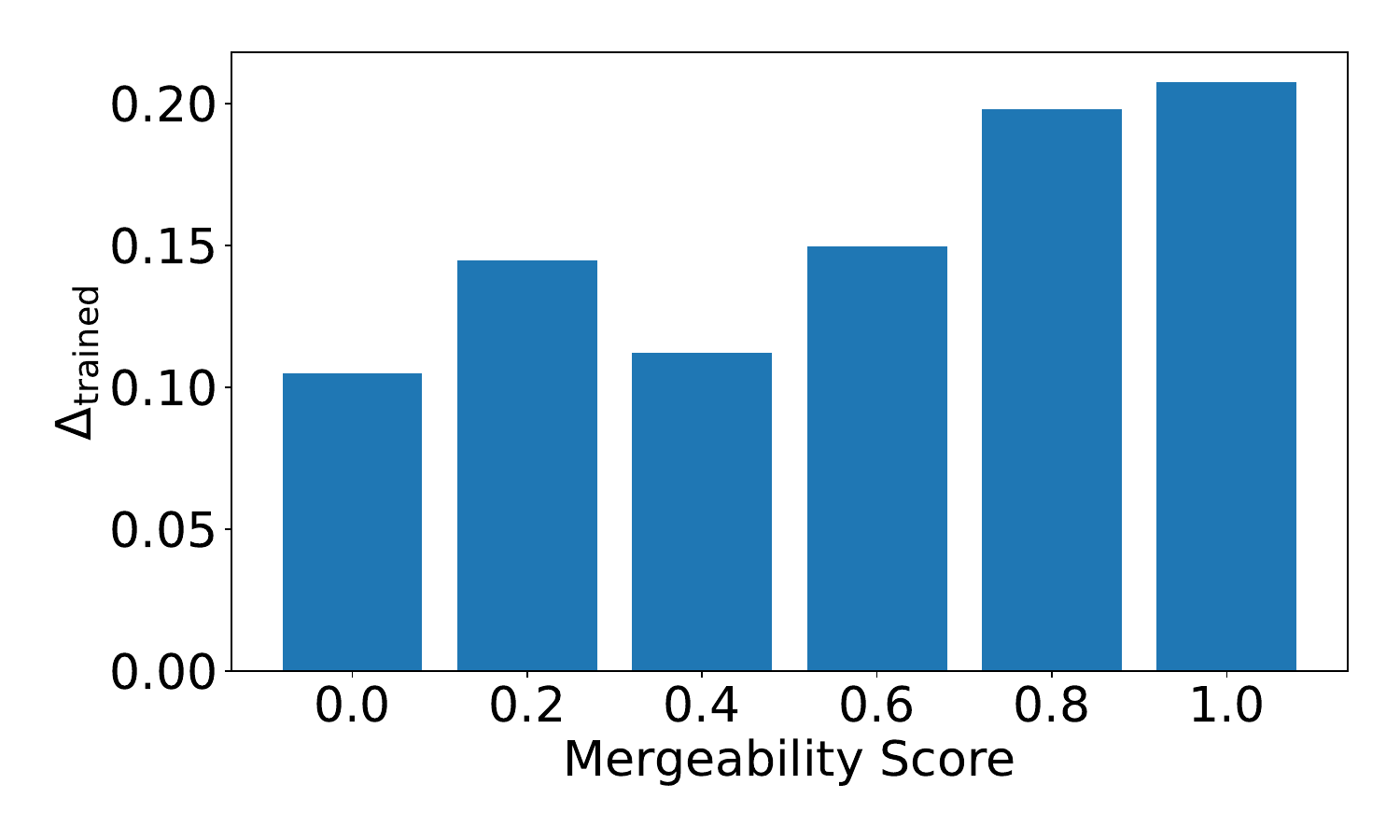}
    \caption{PopQA average difference between post-training and original model answer probability for different mergeability scores. High-mergeability examples achieve larger post-training probability improvements, although they were easier to ``fix''.}
    \label{fig:llama_post_base_dif}
\end{figure}

\begin{figure}
    \centering
    \includegraphics[width=1\linewidth]{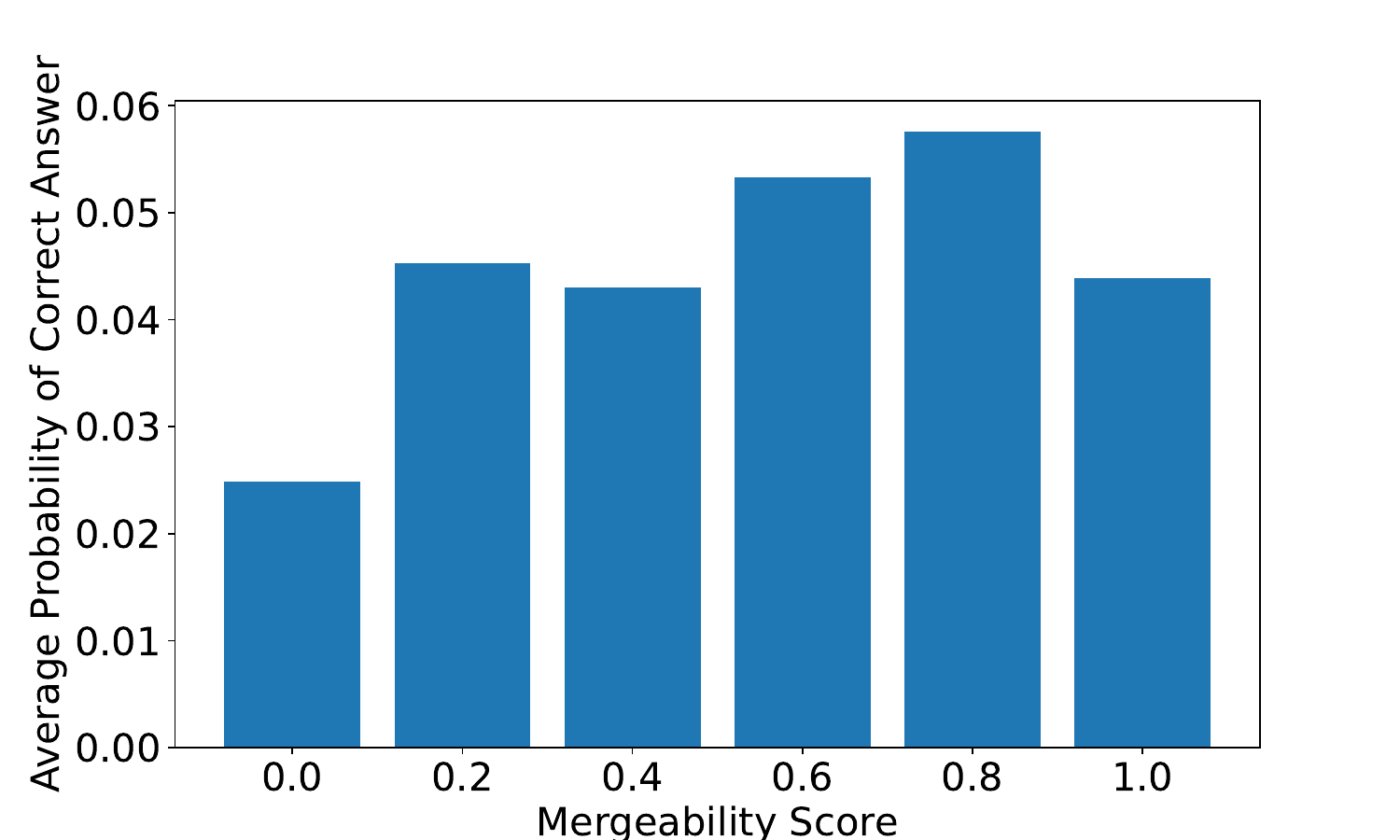}
    \caption{For the PopQA multiple-choice setting, with the Llama model, we report the average probability of the correct answer in the base model.}
    \label{fig:llama_ans_prob}
\end{figure}

\begin{figure}
    \centering
    \includegraphics[width=1\linewidth]{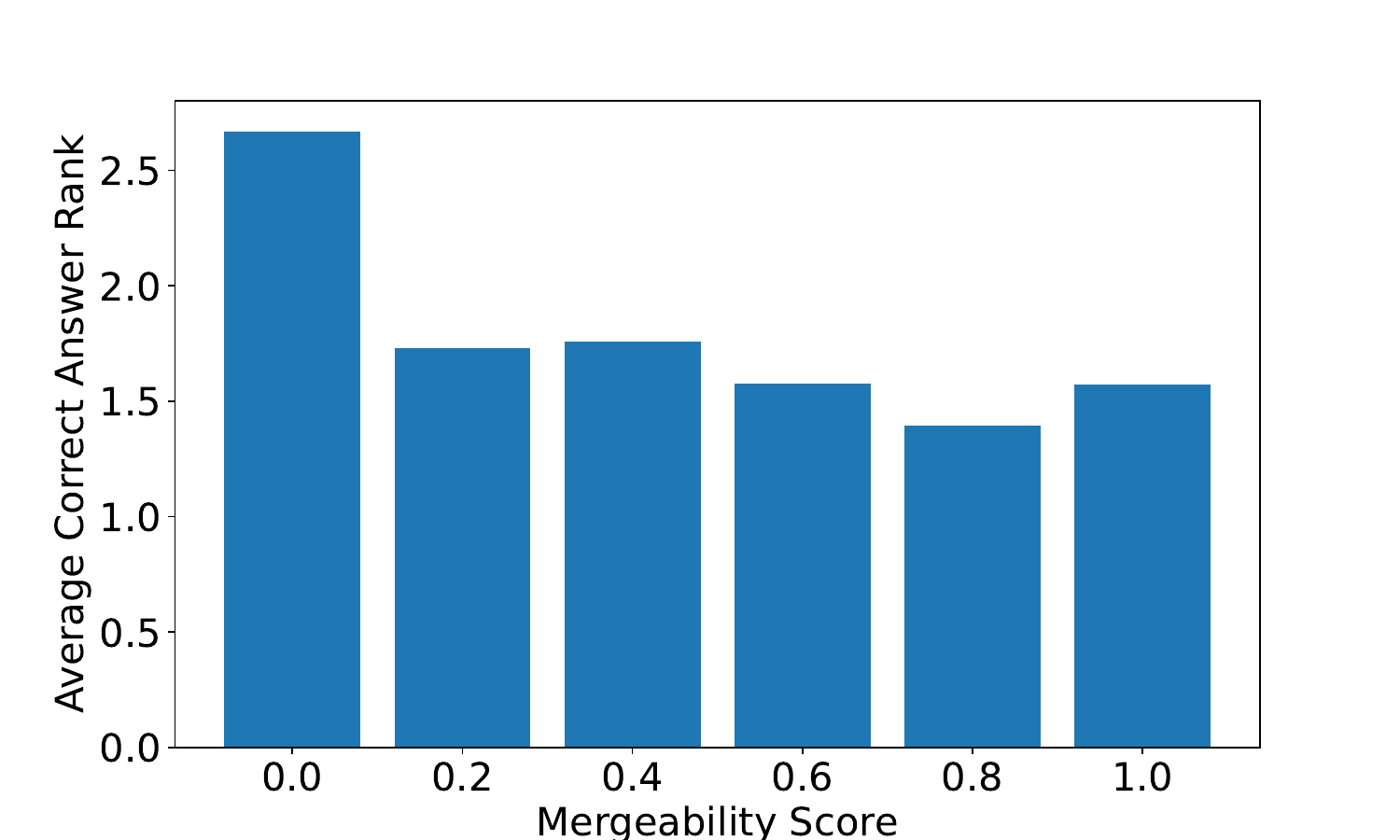}
    \caption{For the PopQA multiple-choice setting, with the Llama model, the figure shows the average rank (lower is better) of the correct answer in the base model of different mergeability scores. While we generally observe that higher-mergeability examples correspond to lower ranks, we also observe a non-monotonic jump for $S=1.0$.}
    \label{fig:llama_answer_rank}
\end{figure}

\begin{figure}
    \centering
    \includegraphics[width=1\linewidth]{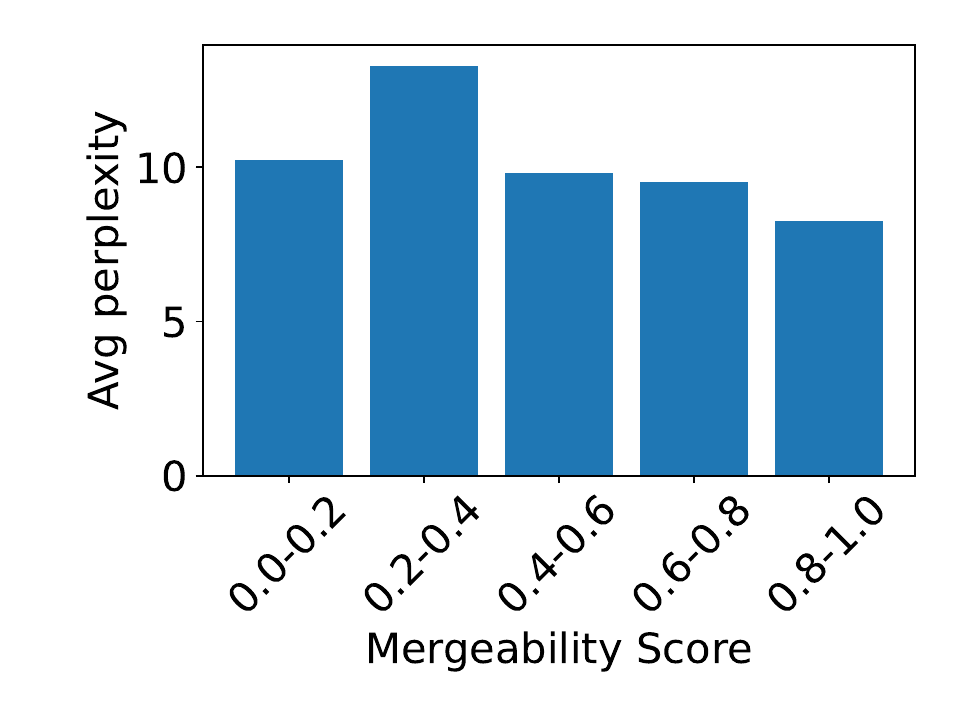}
    \caption{Lots-of-LoRAs train data perplexity of different mergeability scores. We  observe a global trend where higher mergeability examples have a lower average perplexity.}
    \label{fig:lots_of_lora_perplexity_vs_mergeability}
\end{figure}

\begin{figure}[t]
    \centering
    \includegraphics[width=1\linewidth]{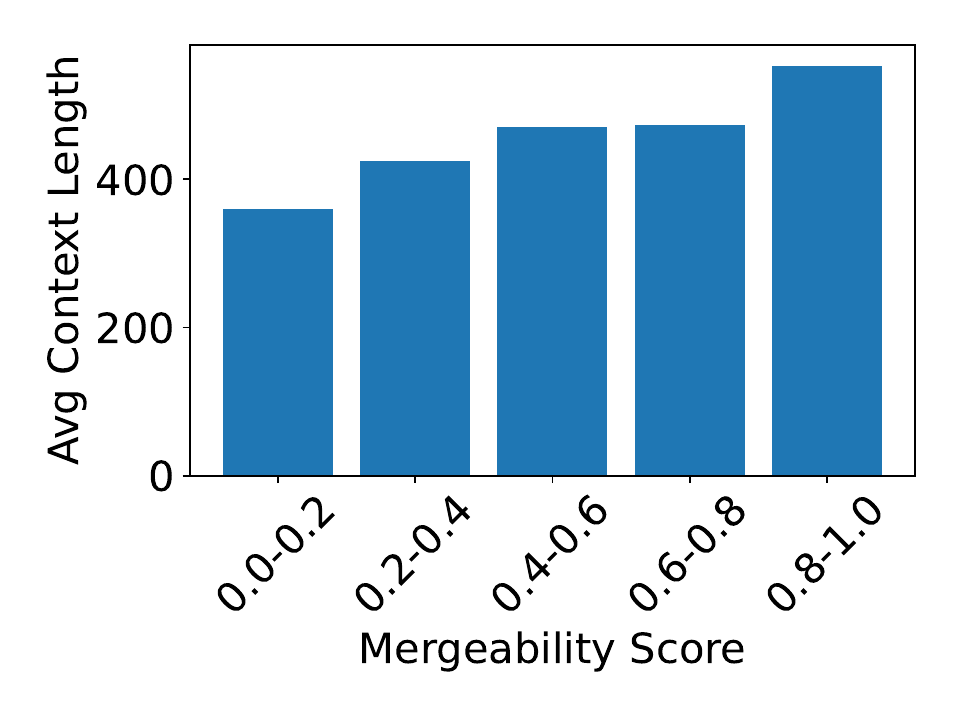}
    \caption{Lots-of-LoRAs train data length (number of tokens) of different mergeability scores. Higher mergeability examples have longer training data on average.}
    \label{fig:lots_of_lora_train_lenght_vs_mergeability}
\end{figure}

In this section, we include additional results of experiments to test the causes of mergeability (\S\ref{sec:mergeability_causes}). 
Figures \ref{fig:llama_highest_correct_diff} and \ref{fig:llama_post_base_dif} show a bar plot of $\Delta_{\text{base}}$ and $\Delta_{\text{trained}}$ results from Figure \ref{fig:popqa_all_results}, respectively. Table \ref{tab:llama_false_results} summarizes the PopQA results for weight properties and general domain knowledge from Figure \ref{fig:popqa_all_results}. Figure \ref{fig:llama_ans_prob} shows the average base model (Llama 3.2) probability of the correct answer. Figure \ref{fig:llama_answer_rank} shows the average rank of the correct answer in the base model (Llama 3.2).
Figures \ref{fig:lots_of_lora_perplexity_vs_mergeability} and \ref{fig:lots_of_lora_train_lenght_vs_mergeability} report the Lots-of-LoRAs experimental result concerning weight properties (\S\ref{sec:rq2}). 


\begin{table*}
\centering
\begin{tabular}{lccccccc@{}}
\toprule
Mergeability score                & 0.0     & 0.2     & 0.4     & 0.6     & 0.8     & 1.0     \\ \midrule
Avg perplexity                           & 9.80  & 9.71  & 9.61  & 9.26  & 9.34  & 10.40 \\
\begin{tabular}[c]{@{}c@{}}Avg context length\\ (\# of tokens)\end{tabular} & 704.36 & 649.25 & 567.84 & 765.77 & 560.33 & 695.59 \\
Avg weight norm                   & 14.91 & 15.11 & 14.90 & 14.81 & 14.67 & 15.28 \\
Avg weight $\sigma_{\text{max}}$ & 18.73 & 19.13 & 18.83 & 18.29 & 18.33 & 19.47 \\ \bottomrule
\end{tabular}
\caption{PopQA results for weight properties and general domain knowledge. There is no clear trend between mergeability score and training data difficulty (average perplexity and average context length) or weight-level properties (average weight norm and average highest singular value).}
\label{tab:llama_false_results}
\end{table*}

\newpage
\subsection{Qwen Experimental Results}

\begin{figure}[h]
    \centering
    \includegraphics[width=1\linewidth]{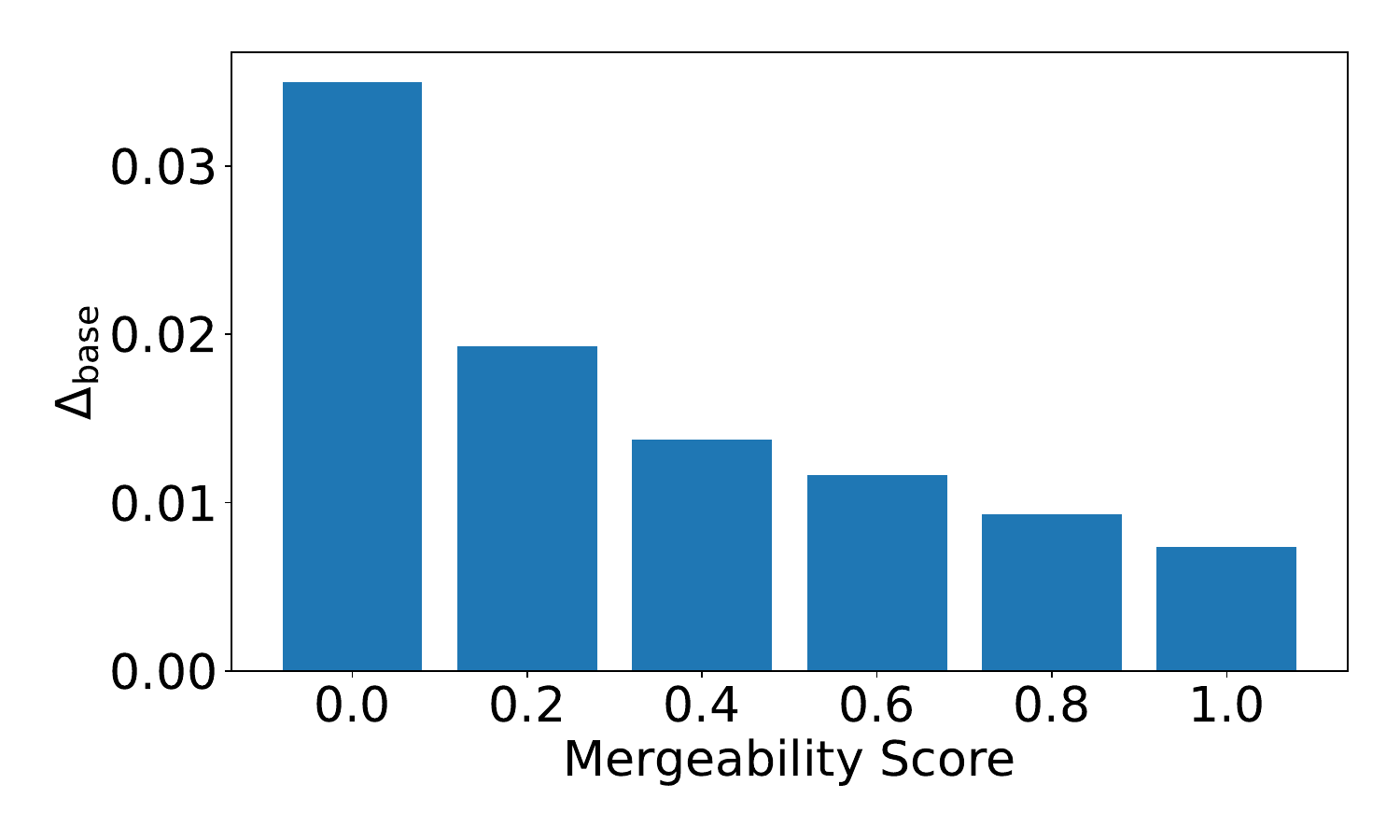}
    \caption{PopQA average difference between the highest and the correct answer probability in the base model (Qwen). We observe that the gap decreases with mergeability, implying that examples with better base model knowledge are more mergeable.}
    \label{fig:qwen_highest_correct_dif}
\end{figure}

Figure \ref{fig:qwen_highest_correct_dif} shows experimental results on PopQA dataset on the Qwen model. Trends are inline with Llama model results (Figure \ref{fig:llama_highest_correct_diff}), with a decrease in the gap as mergeability score increases. This implies that examples with better base-model knowledge are more mergeable.
Table \ref{tab:qwen_results} shows experimental results with the Qwen model for the two other possible causes (\S\ref{sec:rq2}, \S\ref{sec:rq3}). Trends are similar to Llama model results from the main paper (Figure \ref{fig:popqa_all_results}). 

\begin{table*}[]
\begin{tabular}{@{}ccccccc@{}}
\toprule
Mergeability score                & 0.0     & 0.2     & 0.4     & 0.6     & 0.8     & 1.0     \\ \midrule
Avg ppl                           & 12.22 & 11.94 & 10.73 & 14.20 & 10.56 & 12.45 \\
\begin{tabular}[c]{@{}c@{}}Avg context length\\ (\# of tokens)\end{tabular} & 589.52 & 893.25 & 540.48 & 1105.33 & 777.62 & 462.91 \\
Avg weight norm                   & 17.90 & 17.84 & 17.54 & 16.90 & 17.63 & 18.11 \\
Avg weight highest singular value & 26.21 & 25.83 & 25.45 & 23.78 & 25.85 & 26.99 \\
 \bottomrule
\end{tabular}
\caption{Qwen model results for weight properties and training data difficulty in \S\ref{sec:mergeability_causes}.}
\label{tab:qwen_results}
\end{table*}

\subsection{How M and N Values Affect Mergeability?}
\label{sec:additional_distributions}
\begin{figure}[]
    \centering
    \includegraphics[width=0.8\linewidth]{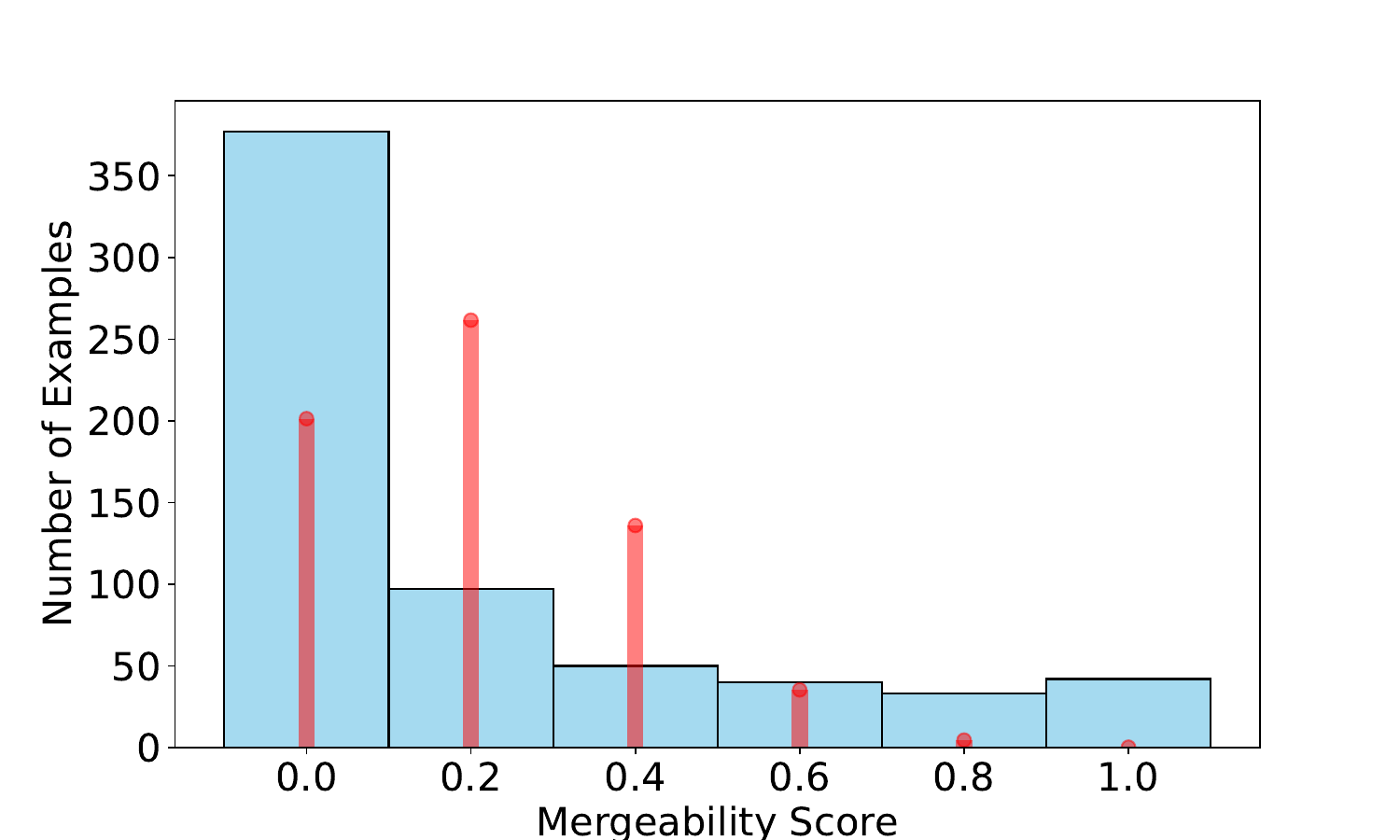}
    \caption{Mergeability scores distribution of Llama-3.2-3B on the PopQA dataset with $N=5, M=50$. Blue and wide bars show the mergeability score as empirically calculated. Red thin bars show the baseline distribution if mergeability was not a model trait, and hence it was a binomial distribution with a fixed success rate.}
    \label{fig:mergeability_score_llama_5}
\end{figure}

\begin{figure}[]
    \centering
    \includegraphics[width=1\linewidth]{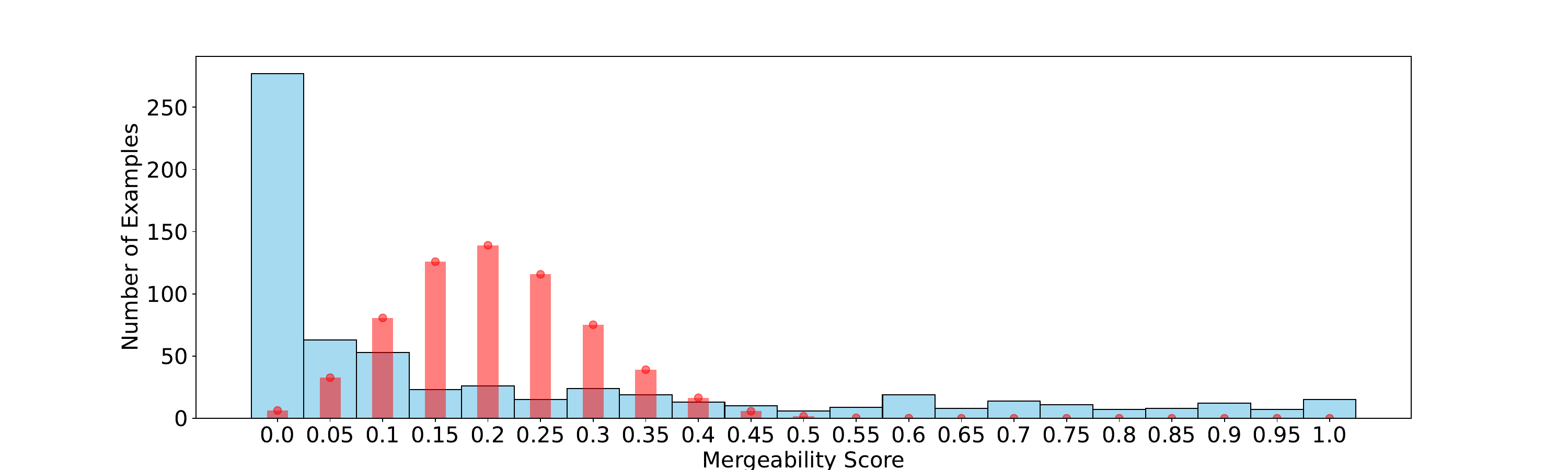}
    \caption{Mergeability scores distribution of Llama-3.2-3B on the PopQA dataset with $N=20, M=50$. Blue and wide bars show the mergeability score as empirically calculated. Red bars show the baseline distribution if mergeability was not a model trait, and hence it was a binomial distribution with a fixed success rate.}
    \label{fig:mergeability_score_llama_20}
\end{figure}

\begin{figure}[]
    \centering
    \includegraphics[width=0.8\linewidth]{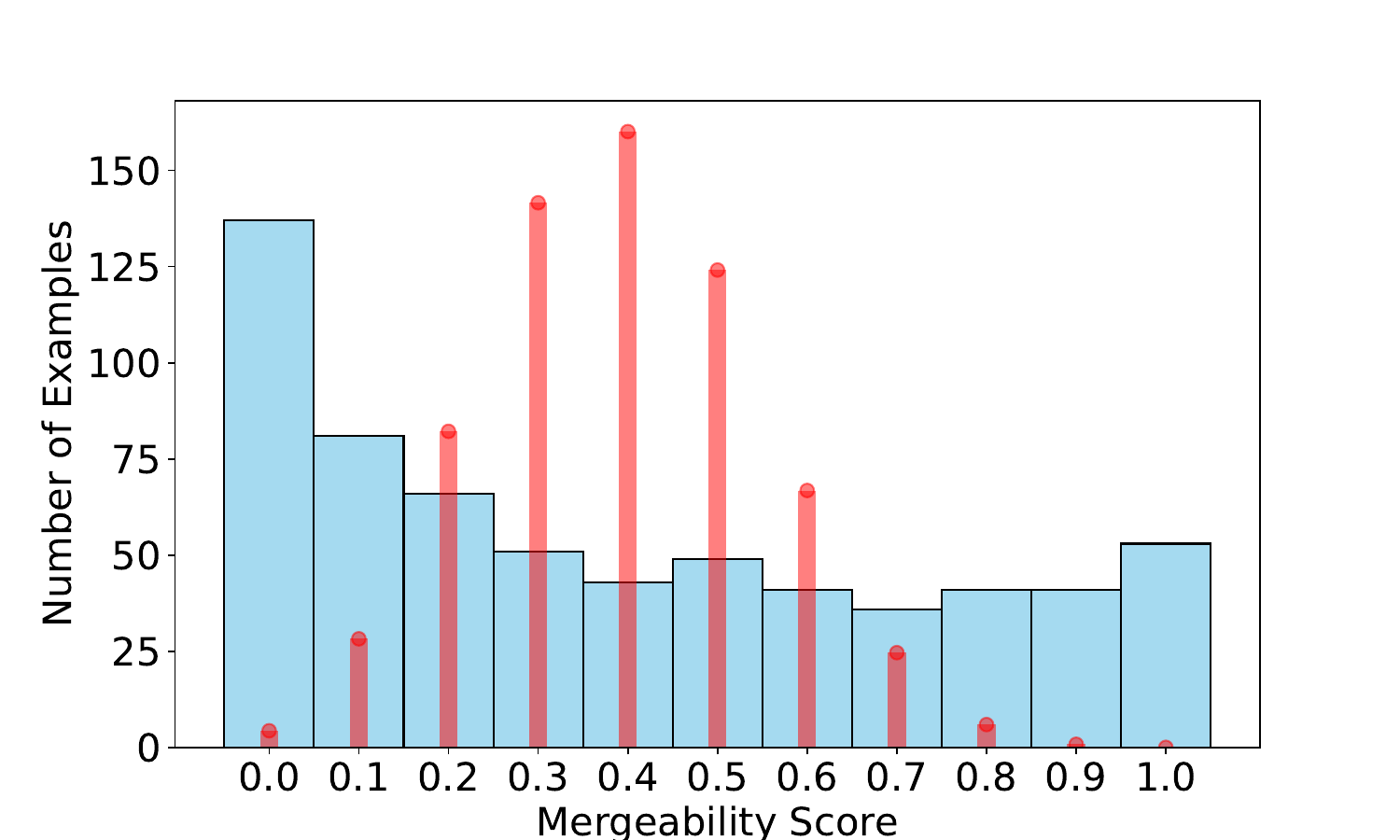}
    \caption{Mergeability scores distribution of Llama-3.2-3B on the PopQA dataset with $N=10, M=10$. Blue and wide bars show the mergeability score as empirically calculated. Red thin bars show the baseline distribution if mergeability was not a model trait, and hence it was a binomial distribution with a fixed success rate. Compared to Figure \ref{fig:llama_mergeability_10}, which shows the results for $M=50$, we see fewer examples with a mergeability score of $0$ and more examples with a higher mergeability score.}
    \label{fig:llama_merge_m=10}
\end{figure}

\begin{figure}[]
    \centering
    \includegraphics[width=1\linewidth]{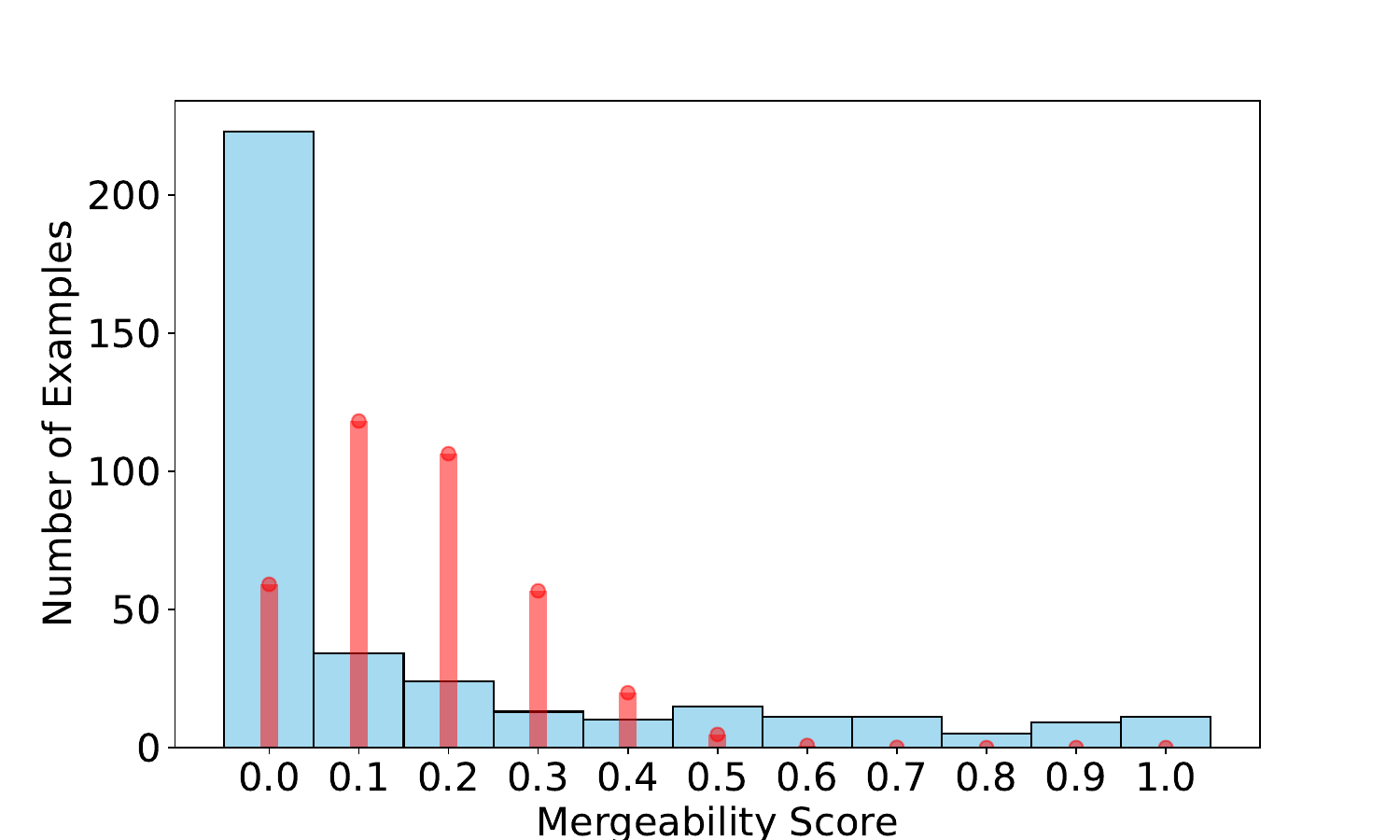}
    \caption{Mergeability scores distribution of Qwen2.5-3B on the PopQA dataset with $N=10, M=50$. Blue and wide bars show the mergeability score as empirically calculated. Red thin bars show the baseline distribution if mergeability was not a model trait, and hence it was a binomial distribution with a fixed success rate.}
    \label{fig:mergeability_score_qwen_10}
\end{figure}

\begin{figure}[]
    \centering
    \includegraphics[width=1\linewidth]{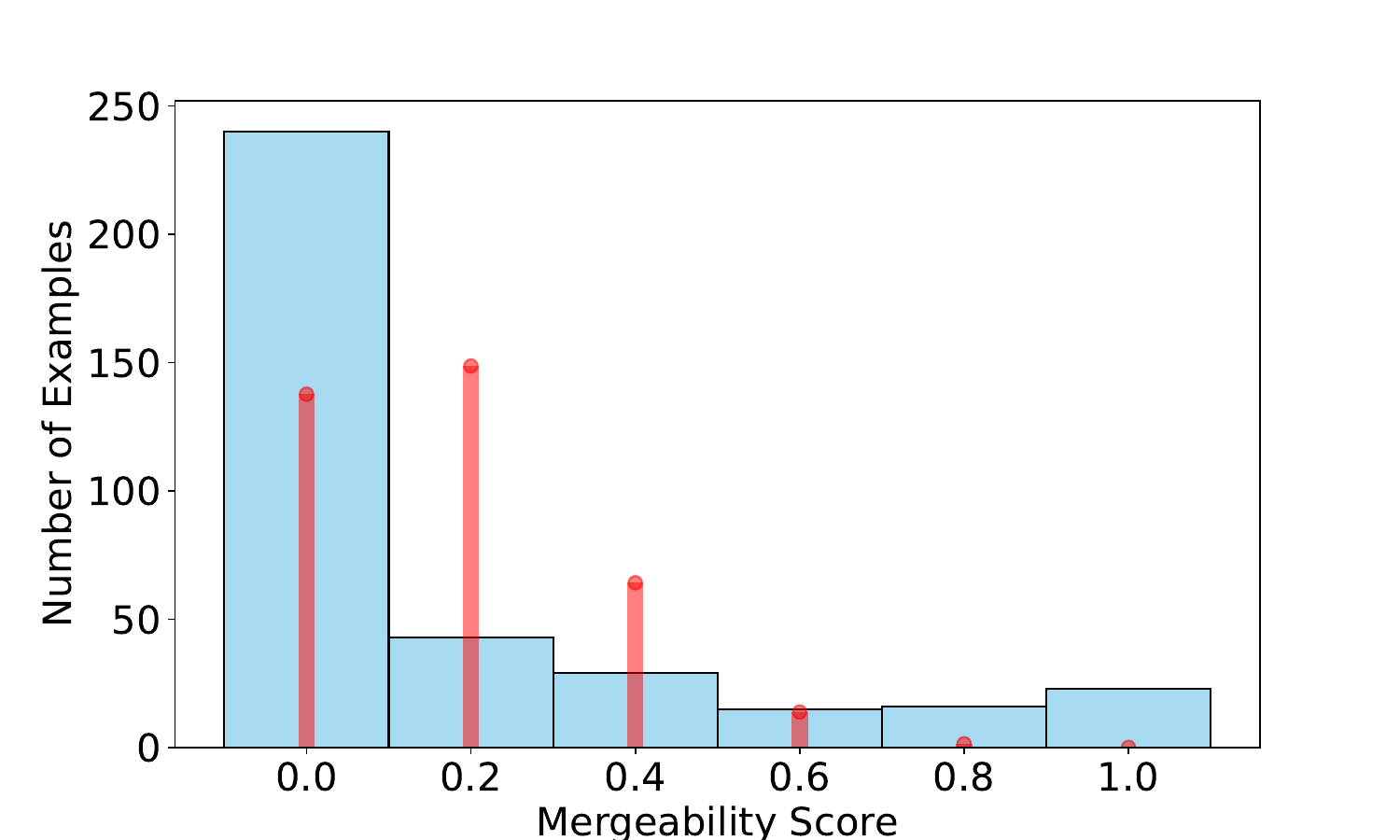}
    \caption{Mergeability scores distribution of Qwen2.5-3B on the PopQA dataset with $N=5, M=50$. Blue and wide bars show the mergeability score as empirically calculated. Red and thin bars show the baseline distribution if mergeability was not a model trait, and hence it was a binomial distribution with a fixed success rate.}
    \label{fig:mergeability_score_qwen_5}
\end{figure}

\begin{figure}[]
    \centering
    \includegraphics[width=1\linewidth]{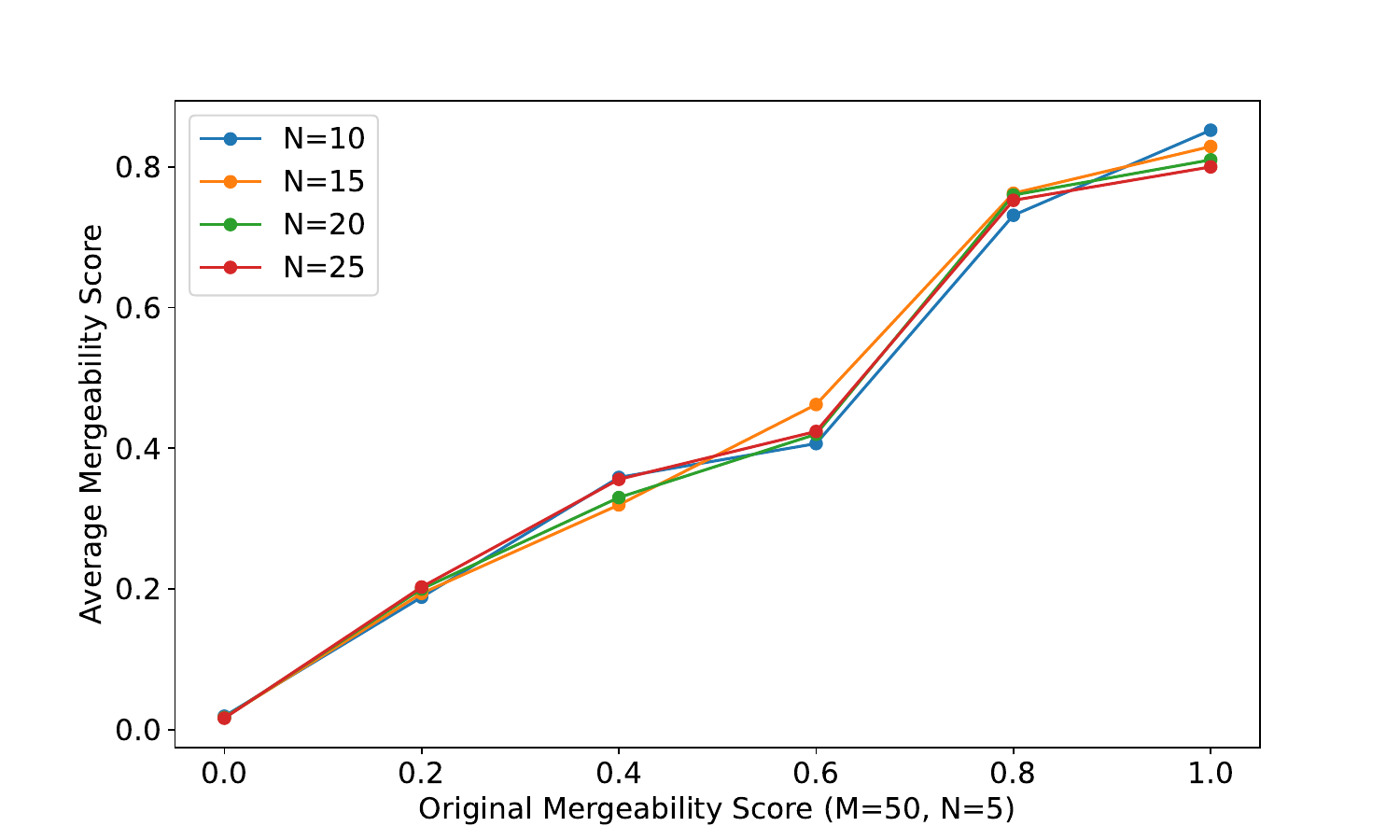}
    \caption{We compare the mergeability score when experimenting with different N values using Qwen2.5-3B on the PopQA dataset. The x-axis is the mergeability score for the $M=50, N=5$ experiment. The y-axis shows the average mergeability score of those examples when experimenting with different $N$ values. We observe an increasing trend for all tested $N$ values, which indicates that examples with a higher mergeability score in $N=5$ also had a higher mergeability score when experimented with other $N$ values.}
    \label{fig:n_values_effect}
\end{figure}

\begin{figure}[]
    \centering
    \includegraphics[width=1\linewidth]{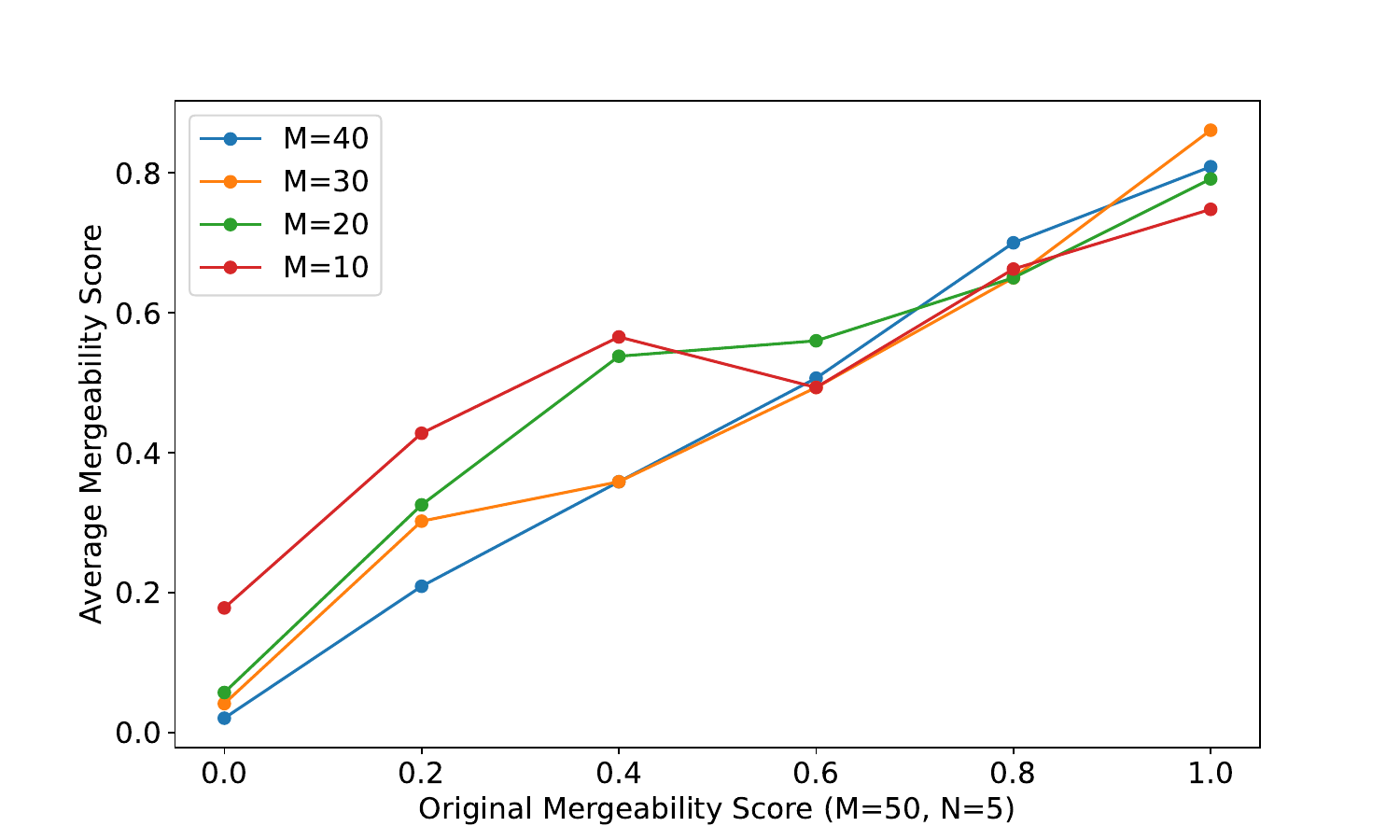}
    \caption{We compare the mergeability score when experimenting with different M values using Qwen2.5-3B on the PopQA dataset. The x-axis is the mergeability score for the $M=50, N=5$ experiment. The y-axis shows the average mergeability score of those examples when experimenting with different $M$ values. We observe a near-perfect increasing trend for all tested $M$ values, which indicates that examples with a higher mergeability score in $M=50$ also had a higher mergeability score when experimented with other $M$ values.}
    \label{fig:m_values_effect}
\end{figure}

Similar to Figure \ref{fig:llama_mergeability_10}, Figure \ref{fig:mergeability_score_llama_5} shows the mergeability distribution of the Llama model with $N=5, M=50$, and Figure \ref{fig:mergeability_score_llama_20} shows the mergeability scores with $N=20, M=50$. We also experiment with different settings for $M$. Figure \ref{fig:llama_merge_m=10} shows the mergeability distribution with $N=5, M=10$. Similar figures for the Qwen model are \ref{fig:mergeability_score_qwen_10} and \ref{fig:mergeability_score_qwen_5} with $N=5, M=50$ and $N=10, <=50$, respectively. All figures show the difference from a baseline distribution and support the existence of mergeability.
We also examined the robustness of the mergeability score to different $N$ and $M$ values. Figures \ref{fig:n_values_effect} and \ref{fig:m_values_effect} show the change in the mergeability score as a function of $N$ and $M$, respectively. In both Figures we see that mergeability scores calculated with one $M, N$ value are correlated with mergeability scores calculated with different $M$ and $N$ values, showing that scores are consistent under different parameters.

\clearpage
\subsection{Lots-of-LoRAs additional experiments}
\label{sec:lots_of_loras_additional}
\begin{figure}[h!]
    \centering
    \includegraphics[width=1\linewidth]{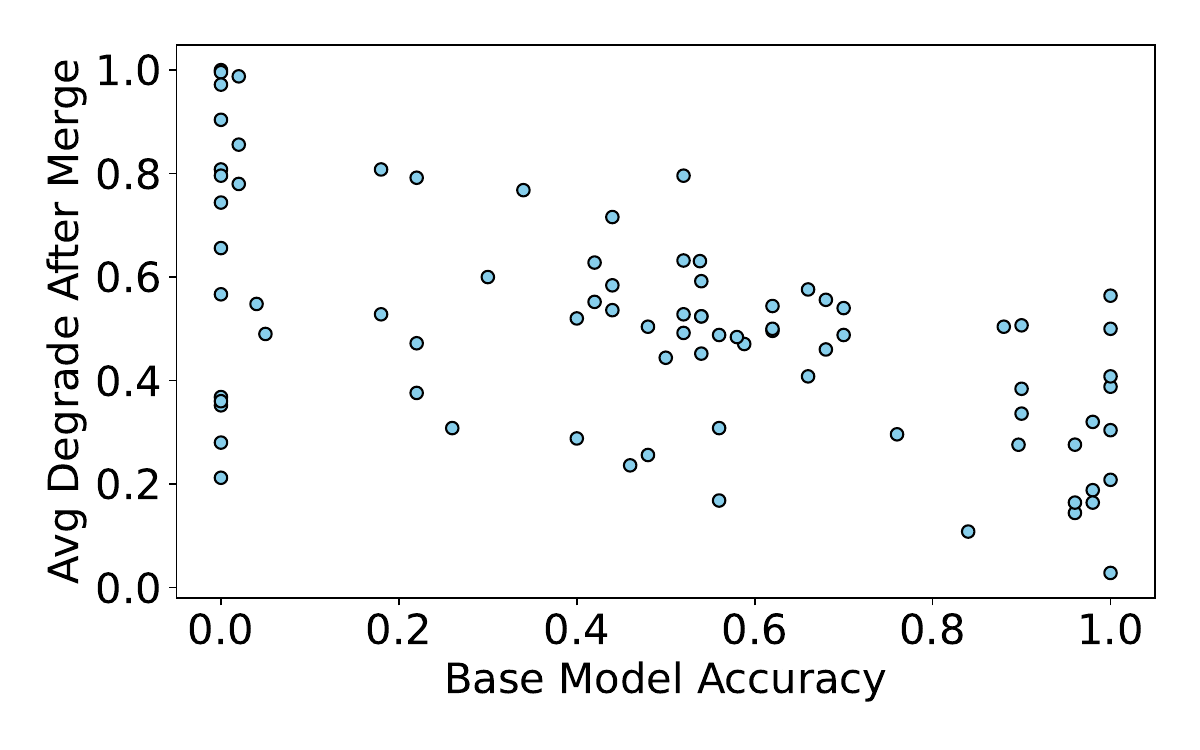}
    \caption{Scatter plot of the average accuracy degradation as a function of the base model accuracy. Figure \ref{fig:lots_of_lora_merge_vs_base_acc} is obtained by calculating the accuracy after degradation and splitting into bins.}
    \label{fig:lota_of_lora_scatter}
\end{figure}

\begin{figure}
    \centering
    \includegraphics[width=1\linewidth]{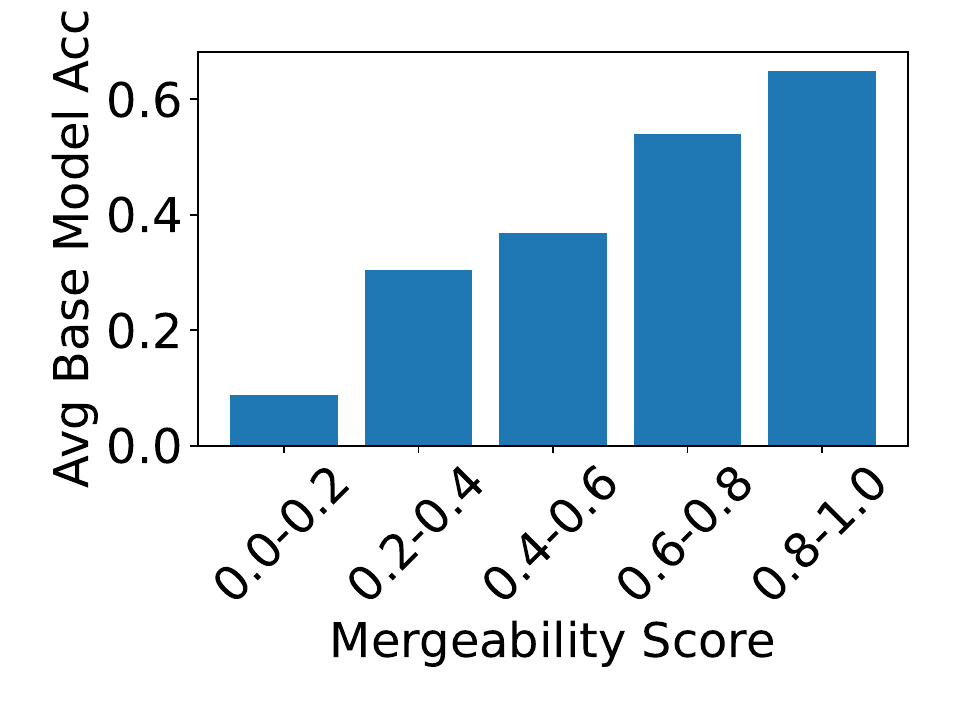}
    \caption{Lots-of-LoRAs average base model task accuracy of different mergeability scores, for tasks with base model accuracy of above 75\%. Higher mergeability scores have on average a higher base model accuracy.}
    \label{fig:lots_of_lora_merge_vs_base_acc_0.75}
\end{figure}

\begin{figure}
    \centering
    \includegraphics[width=1\linewidth]{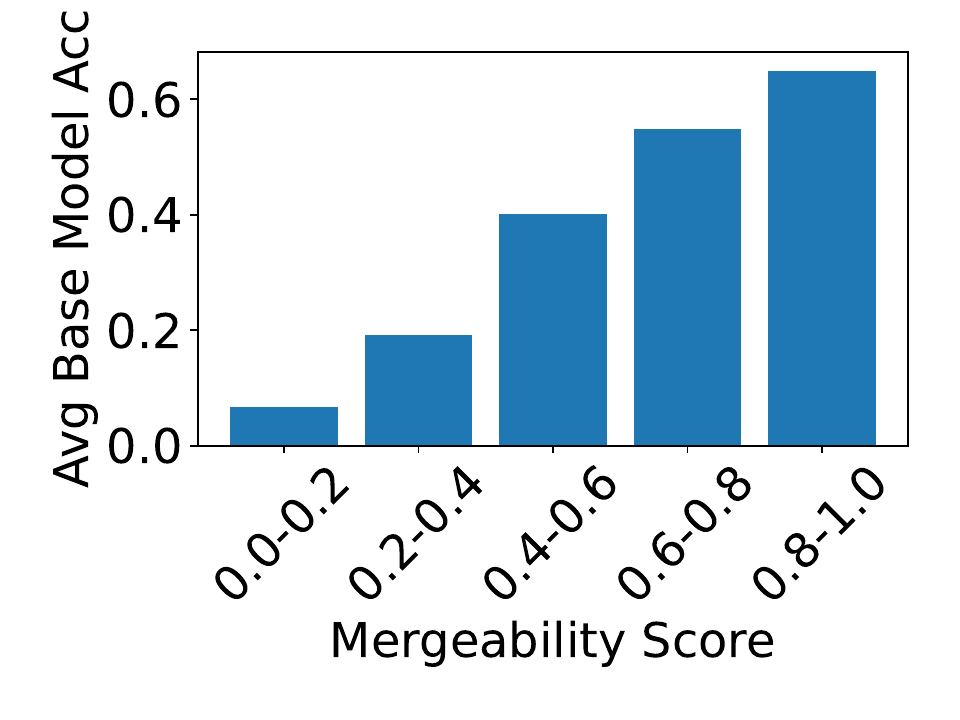}
    \caption{Lots-of-LoRAs average base model task accuracy of different mergeability scores, for tasks with base model accuracy of above 50\%. Higher mergeability scores have on average a higher base model accuracy.}
    \label{fig:lots_of_lora_merge_vs_base_acc_0.5}
\end{figure}

\begin{figure}
    \centering
    \includegraphics[width=1\linewidth]{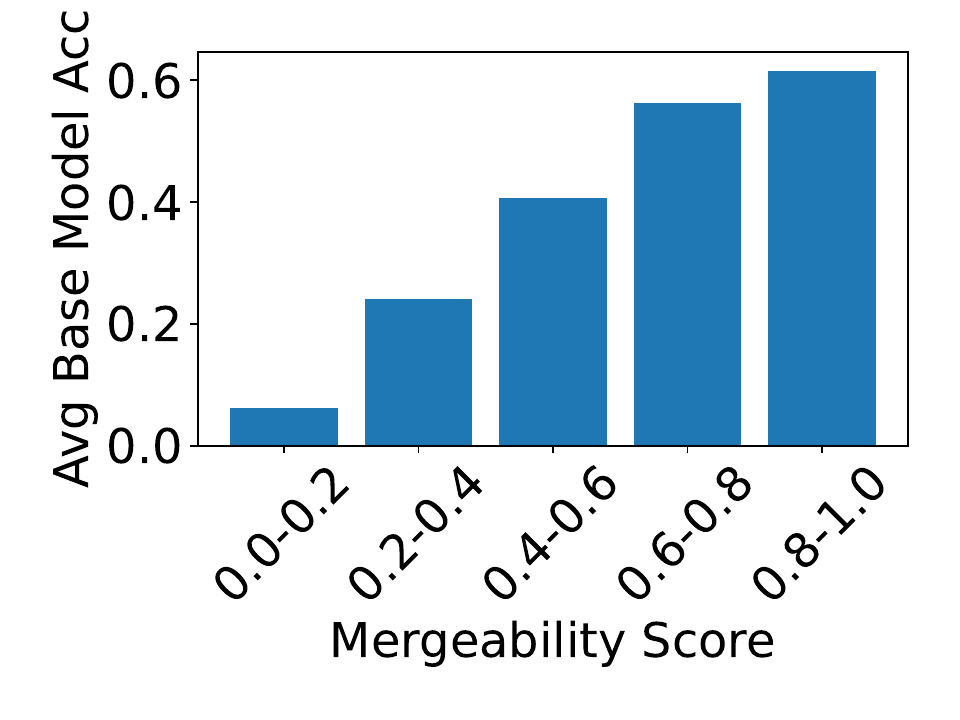}
    \caption{Lots-of-LoRAs average base model task accuracy of different mergeability scores, for tasks with base model accuracy of above 25\%. Higher mergeability scores have on average a higher base model accuracy.}
    \label{fig:lots_of_lora_merge_vs_base_acc_0.25}
\end{figure}

\begin{figure}[t!]
    \centering
    \includegraphics[width=1\linewidth]{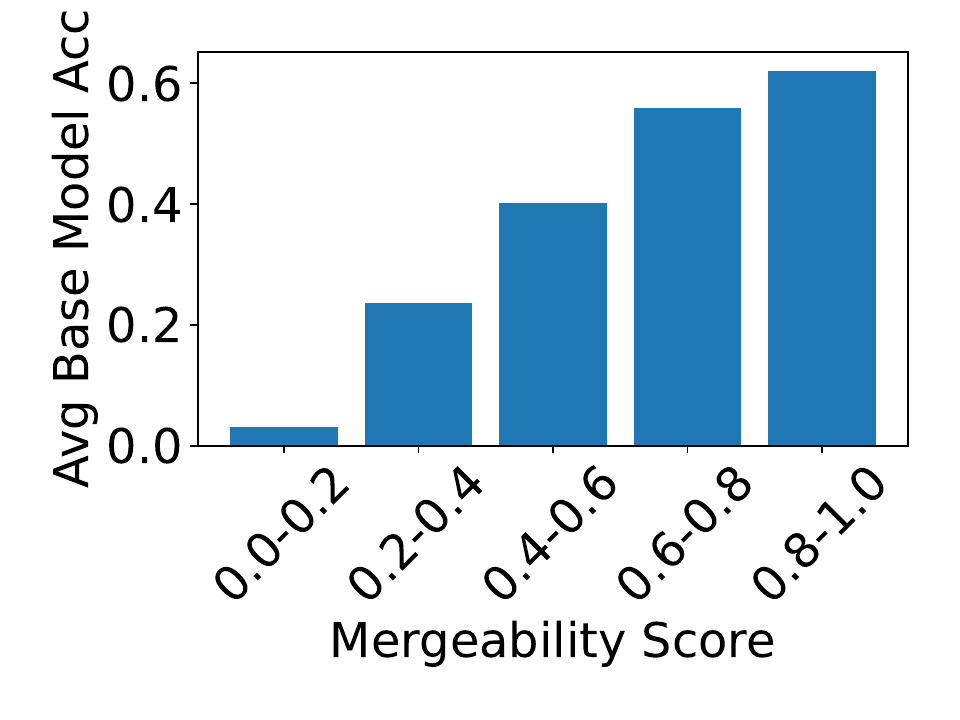}
    \caption{Lots-of-LoRAs average base model task accuracy of different mergeability scores, for all available tasks (tasks with base model accuracy of above 0\%). Higher mergeability scores have on average a higher base model accuracy.}
    \label{fig:lots_of_lora_merge_vs_base_acc_0}
\end{figure}

Figure \ref{fig:lota_of_lora_scatter} shows a scatter plot of the average performance degradation after merging as a function of the base model accuracy. We can observe a general trend of lower degradation for higher base model accuracies. In the main paper (\S\ref{sec:mergeability_causes}), we experimented only with tasks where the finetuned model achieves at least $99\%$ accuracy. In this section we include experimental results with more test accuracy thresholds. Figures \ref{fig:lots_of_lora_merge_vs_base_acc_0.75}, \ref{fig:lots_of_lora_merge_vs_base_acc_0.5}. \ref{fig:lots_of_lora_merge_vs_base_acc_0.25} and \ref{fig:lots_of_lora_merge_vs_base_acc_0} show the average base model accuracy for different mergeability scores when experimenting with tasks that the finetuned model achieves above 75\%, 50\%, 25\% and 0\% accuracy (respectively). We can see that trends are similar across all thresholds (including main paper results at Figure \ref{fig:lots_of_lora_merge_vs_base_acc}), where higher mergeability scores have higher average base model accuracy.

\subsection{Additional Experimental Results}

\begin{figure}[h!]
    \centering
    \includegraphics[width=1\linewidth]{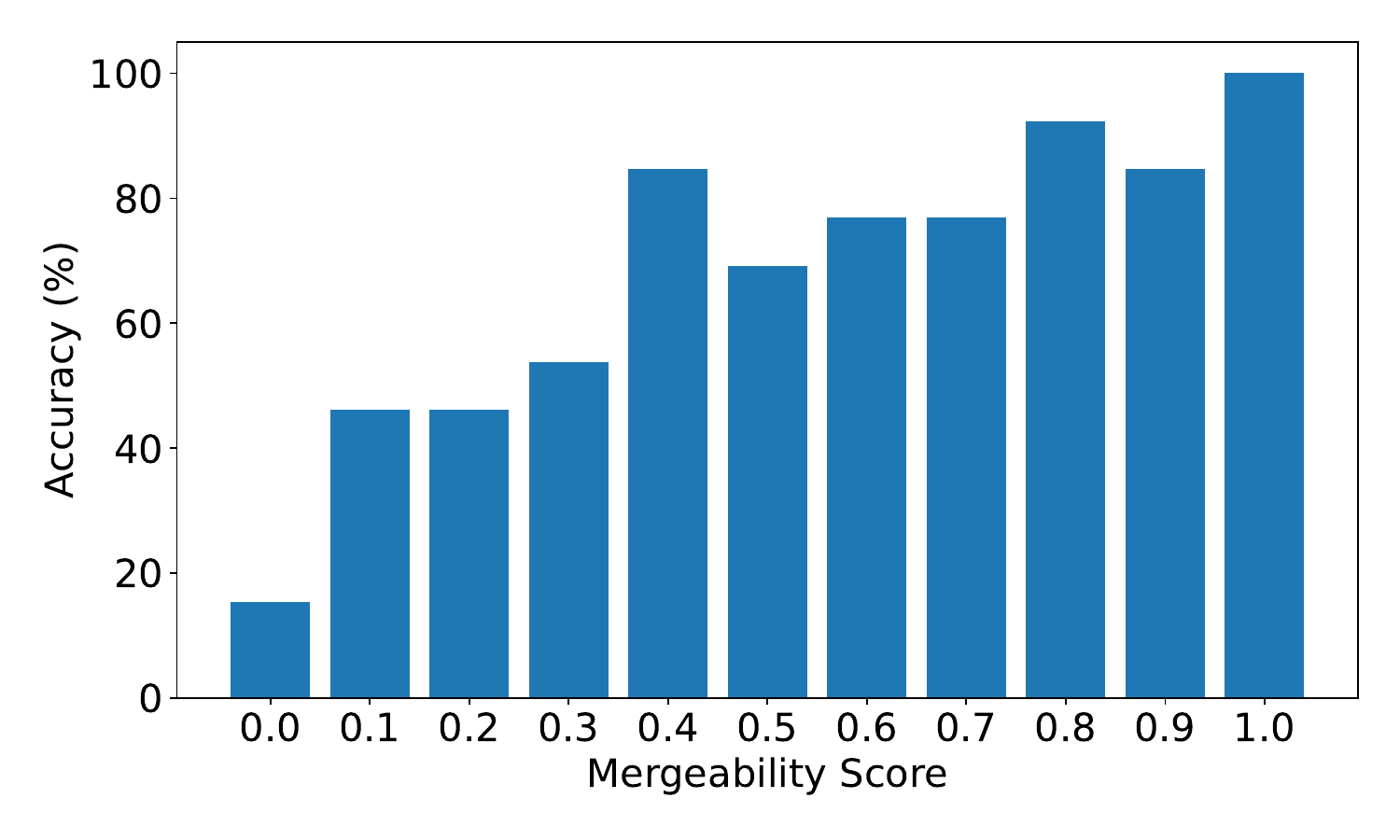}
    \caption{Accuracy when merging examples from the same mergeability group. We can see a trend of higher accuracy as the mergeability score increases.}
    \label{fig:same_score_merging}
\end{figure}

\begin{figure}
\centering
\includegraphics[width=1\linewidth]{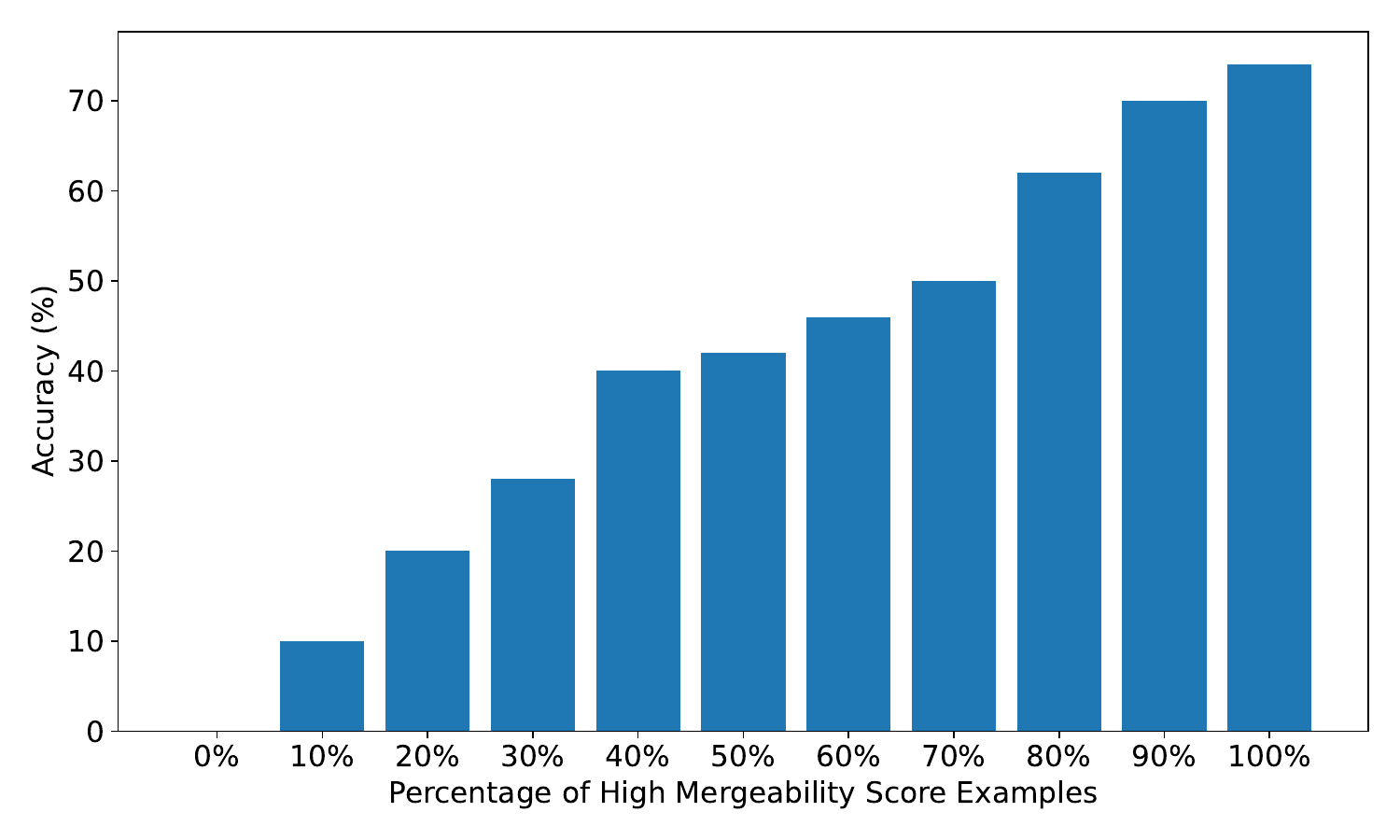}
\caption{Accuracy when merging a set of examples with different percentage of highly mergeable examples. We can see a trend of higher accuracy as more examples are with a high mergeability scores.}
\label{fig:high_mergeable_percentage_change}
\end{figure}

\begin{figure}[]
    \centering
    \includegraphics[width=1\linewidth]{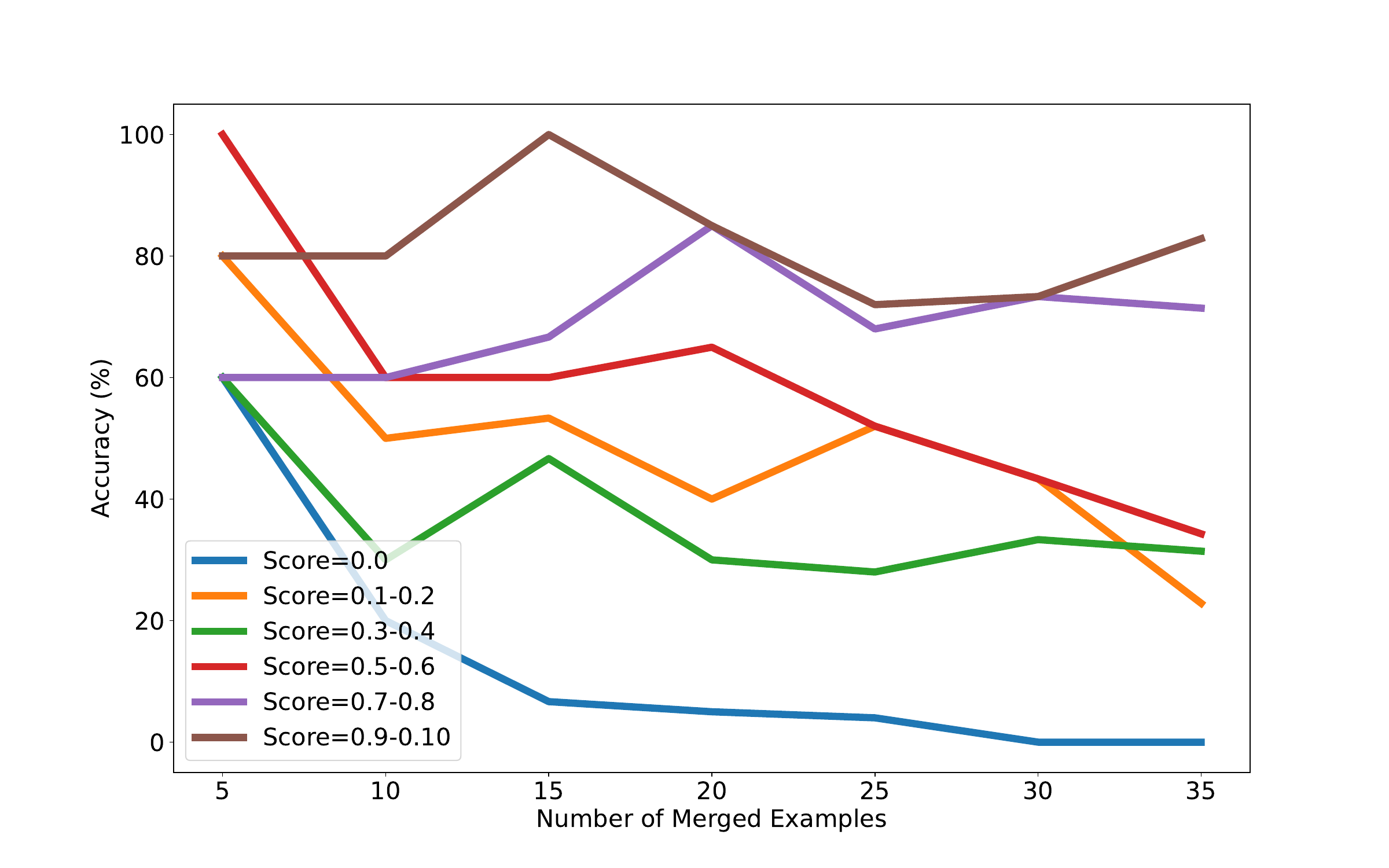}
    \caption{Accuracy for different mergeability scores when merging an increasing number of examples. For a small number of merged examples ($n=5$ for example), we see the lines are not ordered by their mergeability score. As the number of merged examples increases, the lines change their order according to their mergeability score. This shows that the number of merged examples ($M$) might have an effect of the mergeability score.}
    \label{fig:increase_merging}
\end{figure}

Figure \ref{fig:same_score_merging} shows the accuracy when merging examples from a single mergeability group. Higher accuracies are obtained for higher mergeability scores. Figure \ref{fig:high_mergeable_percentage_change} shows the accuracy of a set of examples with different percentages of highly mergeable examples. We took a set of 50 lowest mergeability examples and 50 highest mergeability examples, and each time we merge 50 examples, while changing the percentage of highly mergeable examples from $0\%$ to $100\%$. We see that as the percentage of highly mergeable examples increases, accuracy also increases. Figure \ref{fig:increase_merging} shows the accuracy for different mergeability scores when merging an increasing number of examples. For a small number of merged examples, we see that the lines are not ordered by their mergeability scores. As the number of merged examples increases, the lines change their order according to their mergeability scores. This shows that the number of merged examples ($M$) might have an effect of the mergeability score.

\subsection{Experiments Hyperparameters}
\label{sec:additional_parameters}
For LoRA adapter PopQA training, we use the hyperparameters in Table \ref{tab:popqa_train_param}.
For the mergeability score calculation, we use $M=50$ and $N=5$ or $N=10$ in PopQA, and $M=10$ and $N=5$ for Lots-of-LoRAs.
\begin{table}[h!]
\begin{tabular}{@{}lll@{}}
\toprule
Model                      & Llama        & Qwen         \\ \midrule
LoRA r                     & 64           & 64           \\
LoRA $\alpha$ & 128          & 128          \\
LoRA trained layer         & 14           & 18           \\
LoRA trained param         & mlp.up\_proj & mlp.up\_proj \\
lr                         & 4e-3         & 4e-3         \\
epochs                     & 10           & 10           \\ \bottomrule
\end{tabular}
\caption{PopQA training hyperparameters.}
\label{tab:popqa_train_param}
\end{table}

\subsection{Affect of LoRA Rank}
\label{sec:lora_rank_affect}
\begin{figure}
    \centering
    \includegraphics[width=1\linewidth]{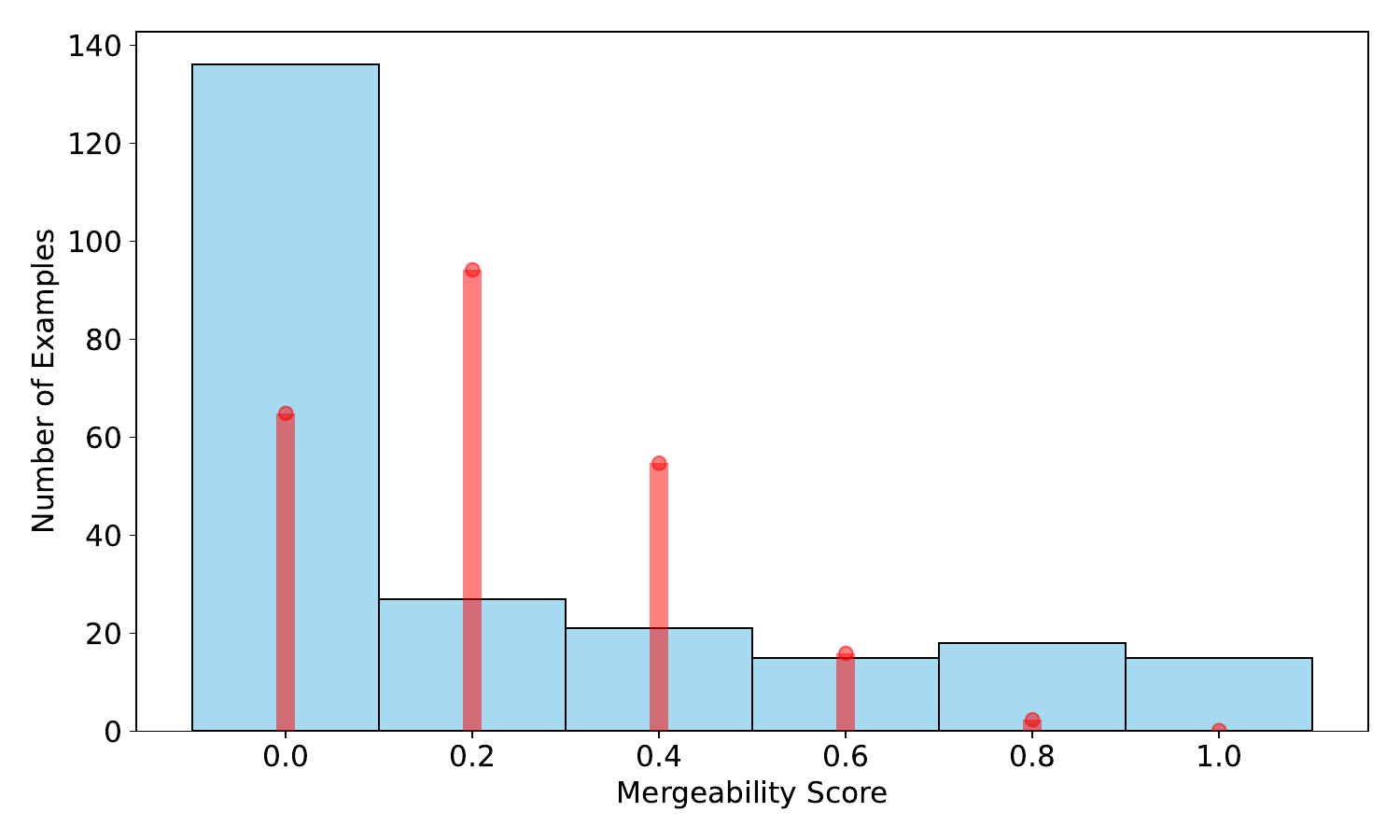}
    \caption{Mergeability scores distribution of Qwen2.5-3B on the PopQA dataset with $N=5, M=50$ and LoRA rank $r=8$. Blue and wide bars show the mergeability score as empirically calculated. Red and thin bars show the baseline distribution if mergeability was not a model trait, and hence it was a binomial distribution with a fixed success rate.}
    \label{fig:qwen_r8_histogram}
\end{figure}

\begin{figure}
    \centering
    \includegraphics[width=1\linewidth]{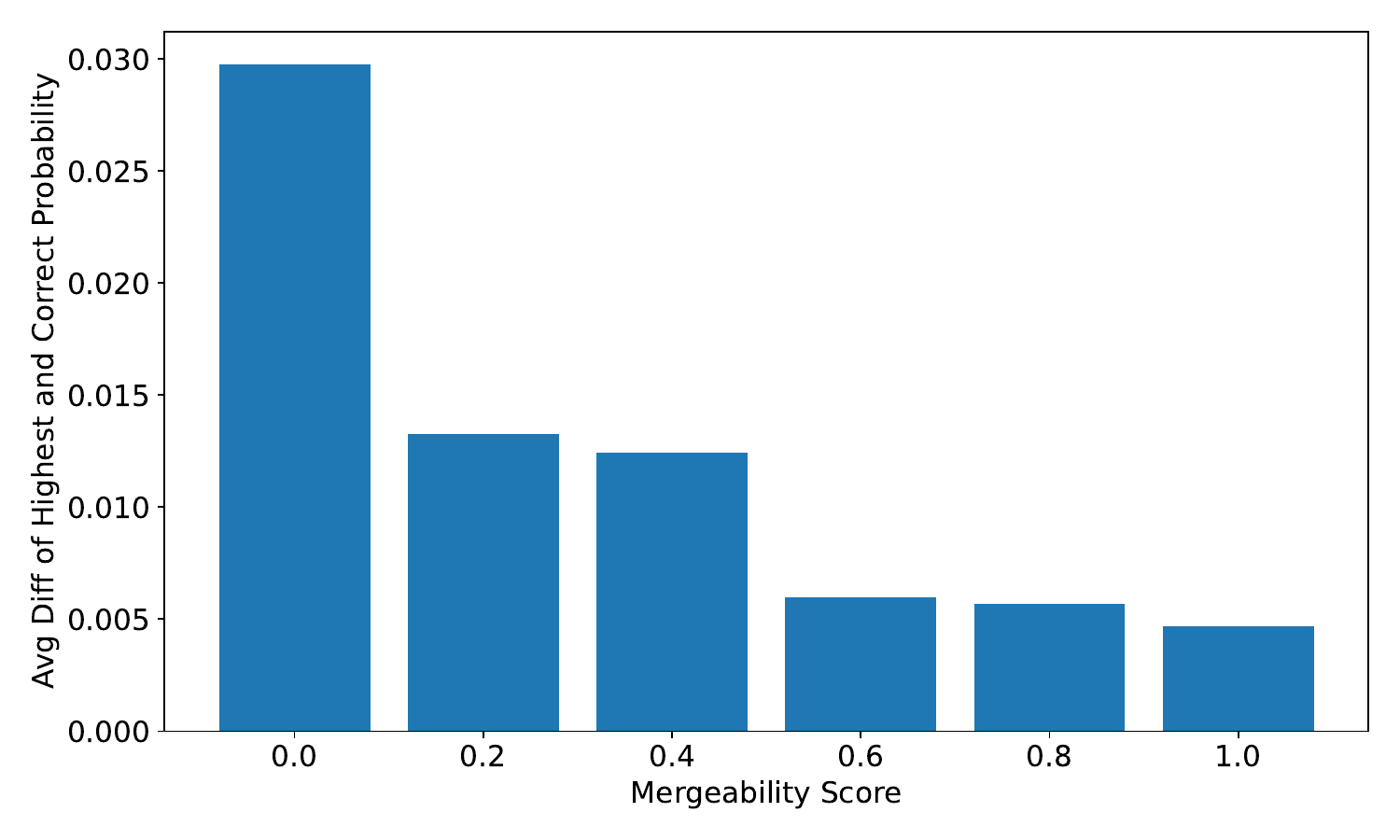}
    \caption{PopQA average difference between the highest and the correct answer probability in the base model (Qwen). Mergeability scores are for examples trained with LoRA rank $r=8$. We observe that the gap decreases with mergeability, implying that examples with better base model knowledge are more mergeable.}
    \label{fig:qwen_r8_results}
\end{figure}

\begin{figure}
    \centering
    \includegraphics[width=1\linewidth]{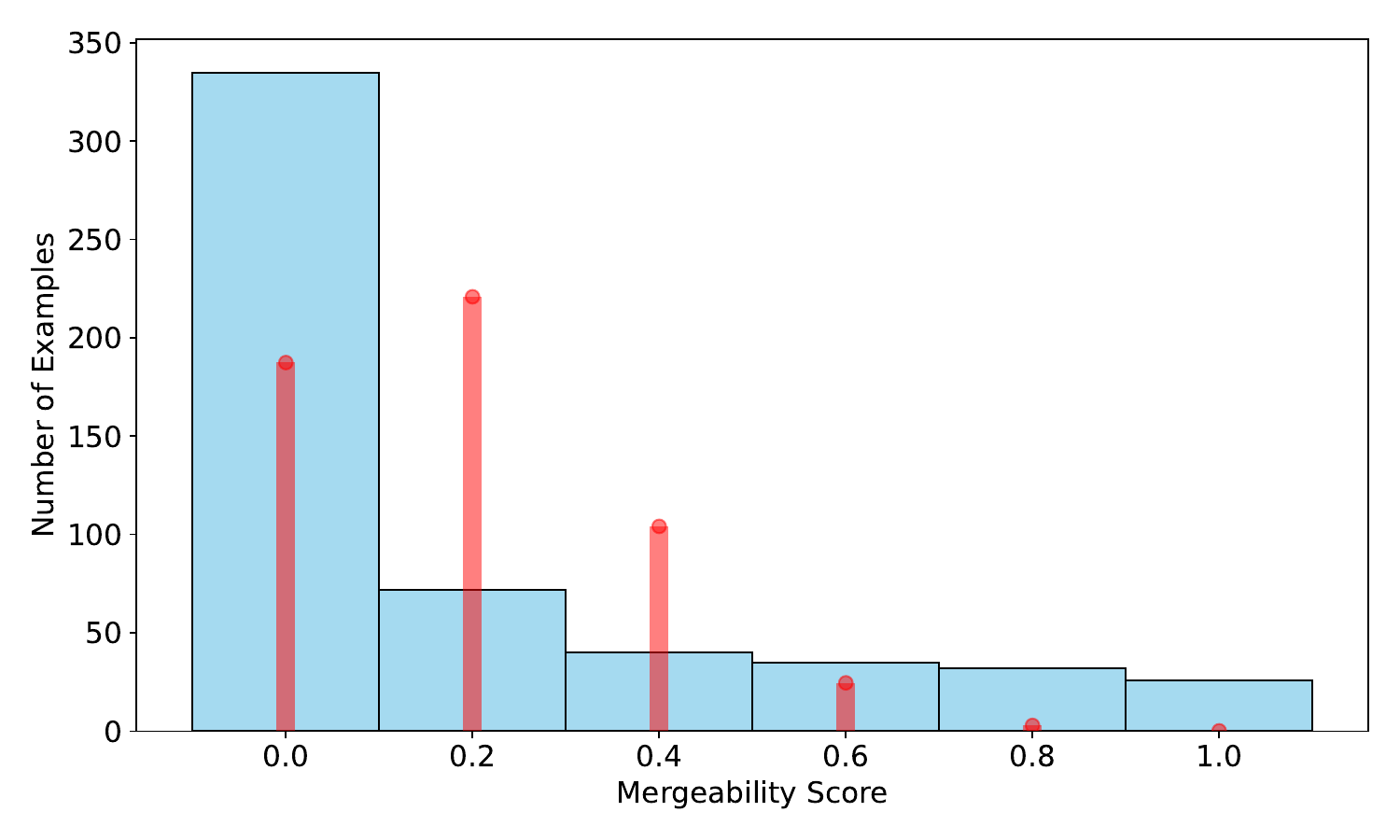}
    \caption{Mergeability scores distribution of Qwen2.5-3B on the PopQA dataset with $N=5, M=50$ and LoRA rank $r=256$. Blue and wide bars show the mergeability score as empirically calculated. Red and thin bars show the baseline distribution if mergability was not a model trait, and hence it was a binomial distribution with a fixed success rate.}
    \label{fig:qwen_r256_histogram}
\end{figure}

\begin{figure}
    \centering
    \includegraphics[width=1\linewidth]{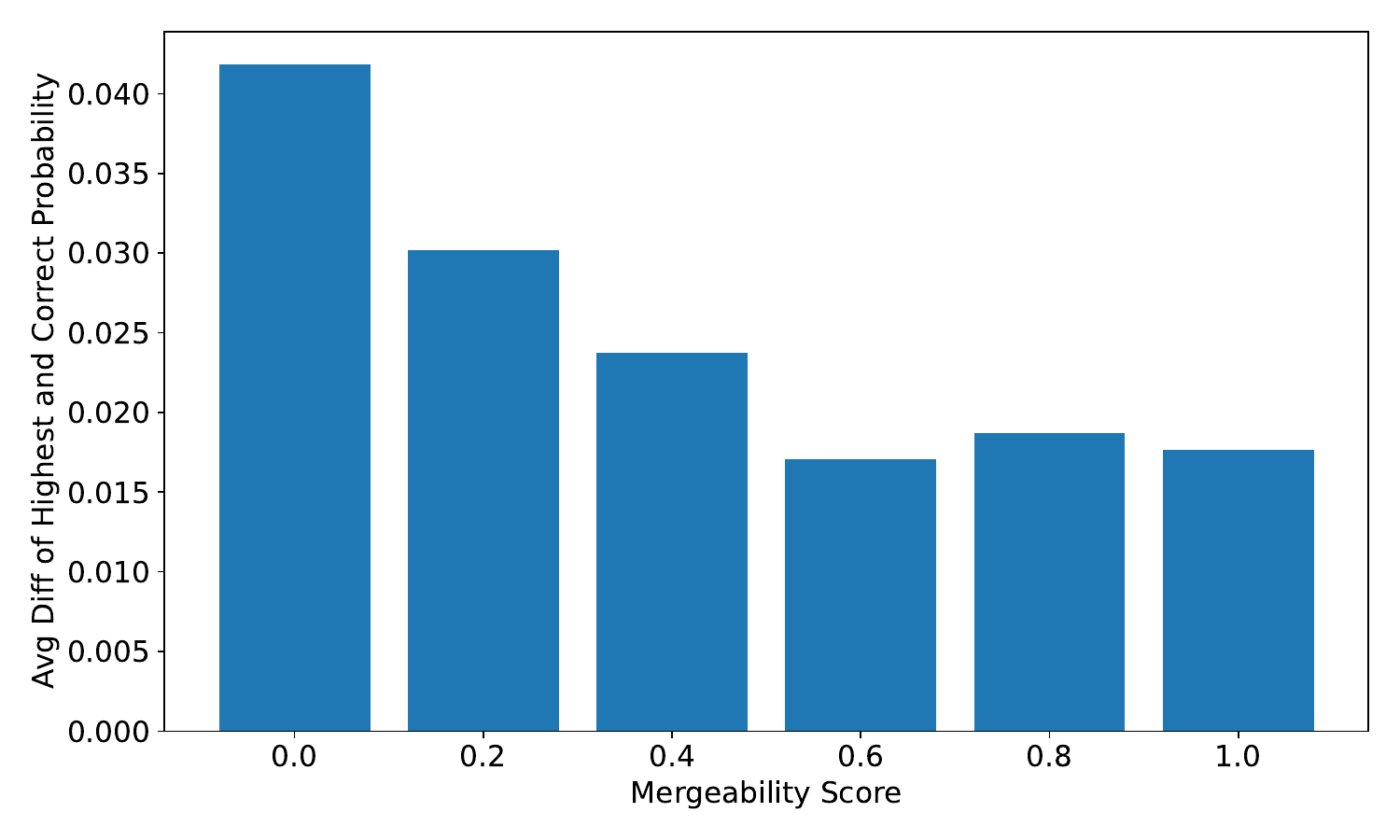}
    \caption{PopQA average difference between the highest and the correct answer probability in the base model (Qwen). Mergeability scores are for examples trained with LoRA rank $r=256$. We observe that the gap generally decreases with mergeability, implying that examples with better base model knowledge are more mergeable.}
    \label{fig:qwen_r256_results}
\end{figure}

In the main text (\S\ref{sec:mergeability_causes}) we experimented with a fixed LoRA rank $r=64$. In this section, we include results for additional LoRA ranks, $r=8$ and $r=256$.
Besides the change of the rank, we maintained the identical experimental setup described in \S\ref{sec:setup}, using the Qwen 3B base model and evaluating using the example-level PopQA dataset. Figures \ref{fig:qwen_r8_histogram}, \ref{fig:qwen_r256_histogram} show the mergeability score distribution for $r=8/256$, respectively. Both figures show the difference from a baseline distribution and support the existence of mergeability. Figures \ref{fig:qwen_r8_results}, \ref{fig:qwen_r256_results} show experimental results for $r=8/256$, respectively. For $r=8$ trends are inline with $r=64$ results (Figure \ref{fig:qwen_highest_correct_dif}), with lower differences for higher mergeability scores. For $r=256$ we also observe a general decreasing effect. However, trends are less sharp compared to $r=64$ results.

\subsection{Mergeability of Full Finetuning}
\label{sec:full_finetuning}
\begin{figure}
    \centering
    \includegraphics[width=1\linewidth]{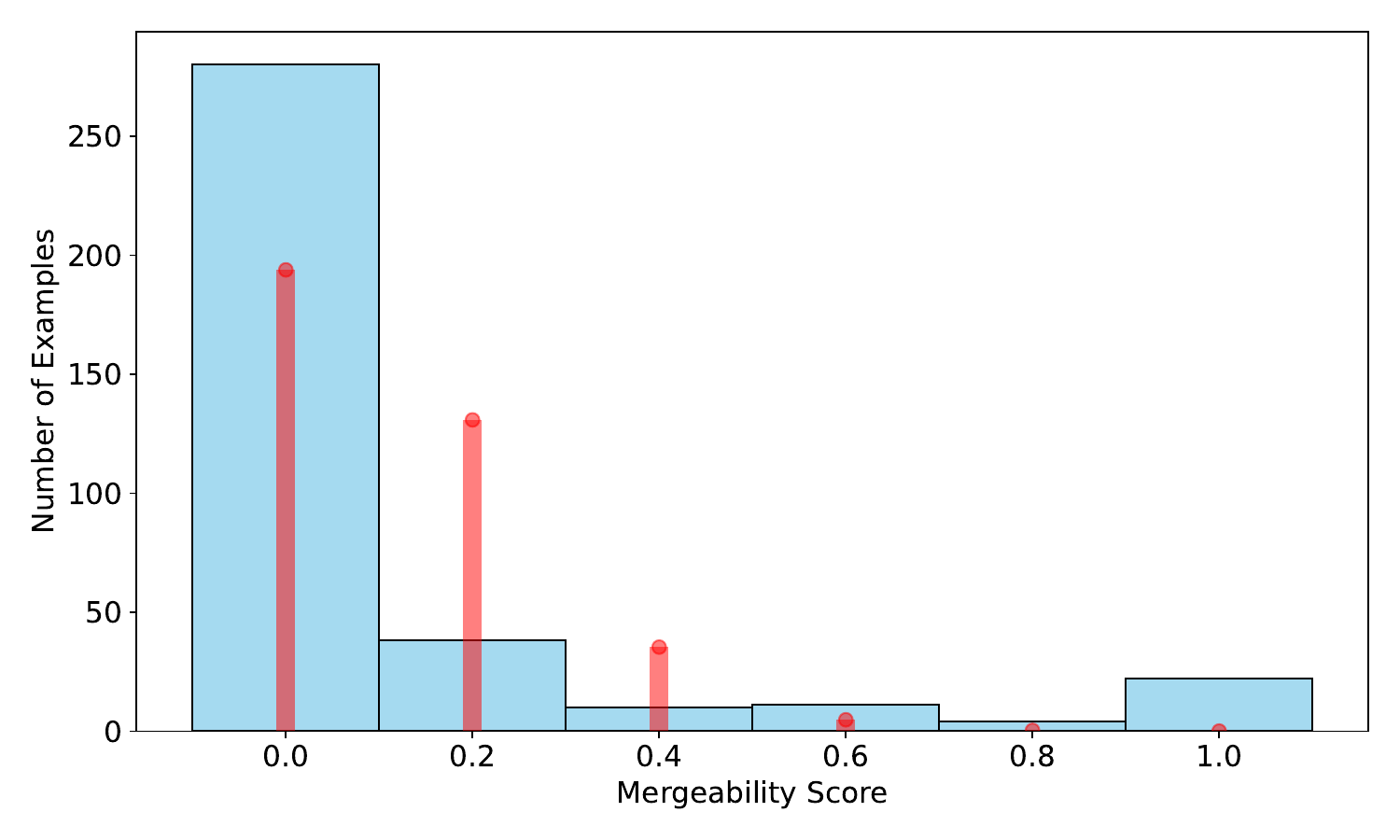}
    \caption{Mergeability scores distribution of Qwen2.5-3B on the PopQA dataset with $N=5, M=50$ and full finetuning training. The scores were calculated using mean merging. Blue and wide bars show the mergeability score as empirically calculated. Red and thin bars show the baseline distribution if mergeability was not a model trait, and hence it was a binomial distribution with a fixed success rate.}
    \label{fig:full_ft_mean_histogram}
\end{figure}

\begin{figure}
    \centering
    \includegraphics[width=1\linewidth]{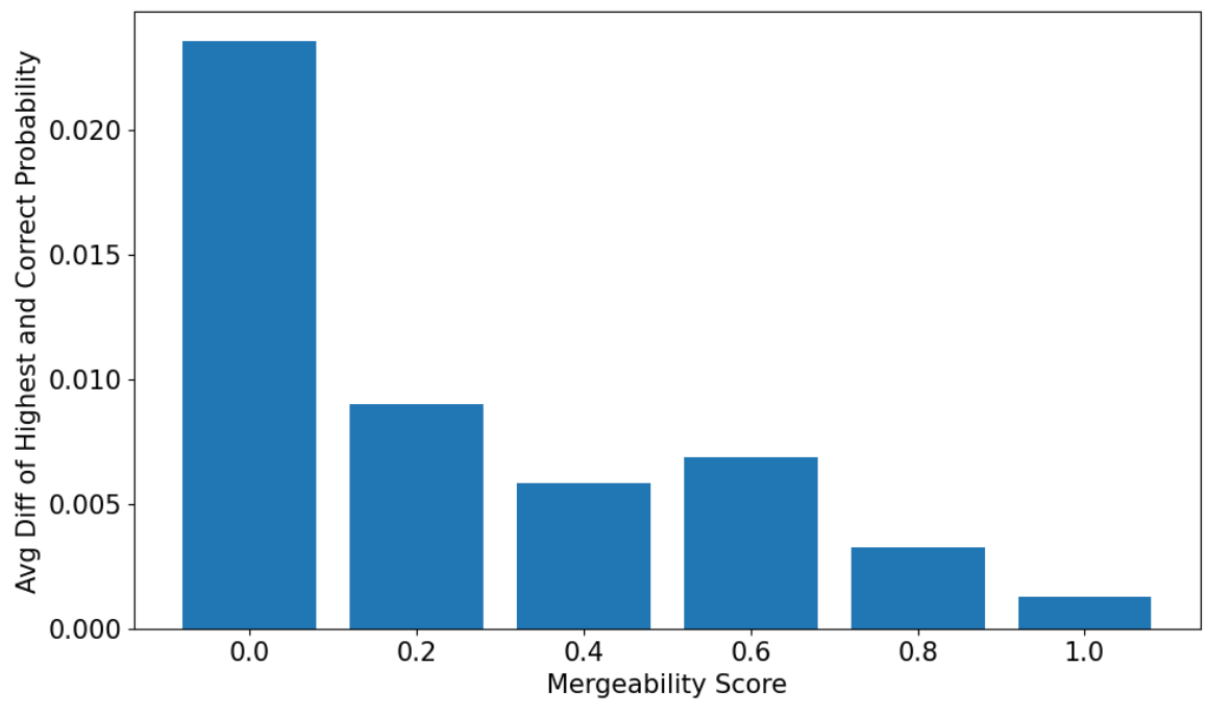}
    \caption{PopQA average difference between the highest and the correct answer probability in the base model (Qwen). Mergeability scores are for examples trained with full finetuning. The scores were calculated using mean merging. We observe that the gap generally decreases with mergeability, implying that examples with better base model knowledge are more mergeable.}
    \label{fig:full_ft_mean_results}
\end{figure}

\begin{figure}
    \centering
    \includegraphics[width=1\linewidth]{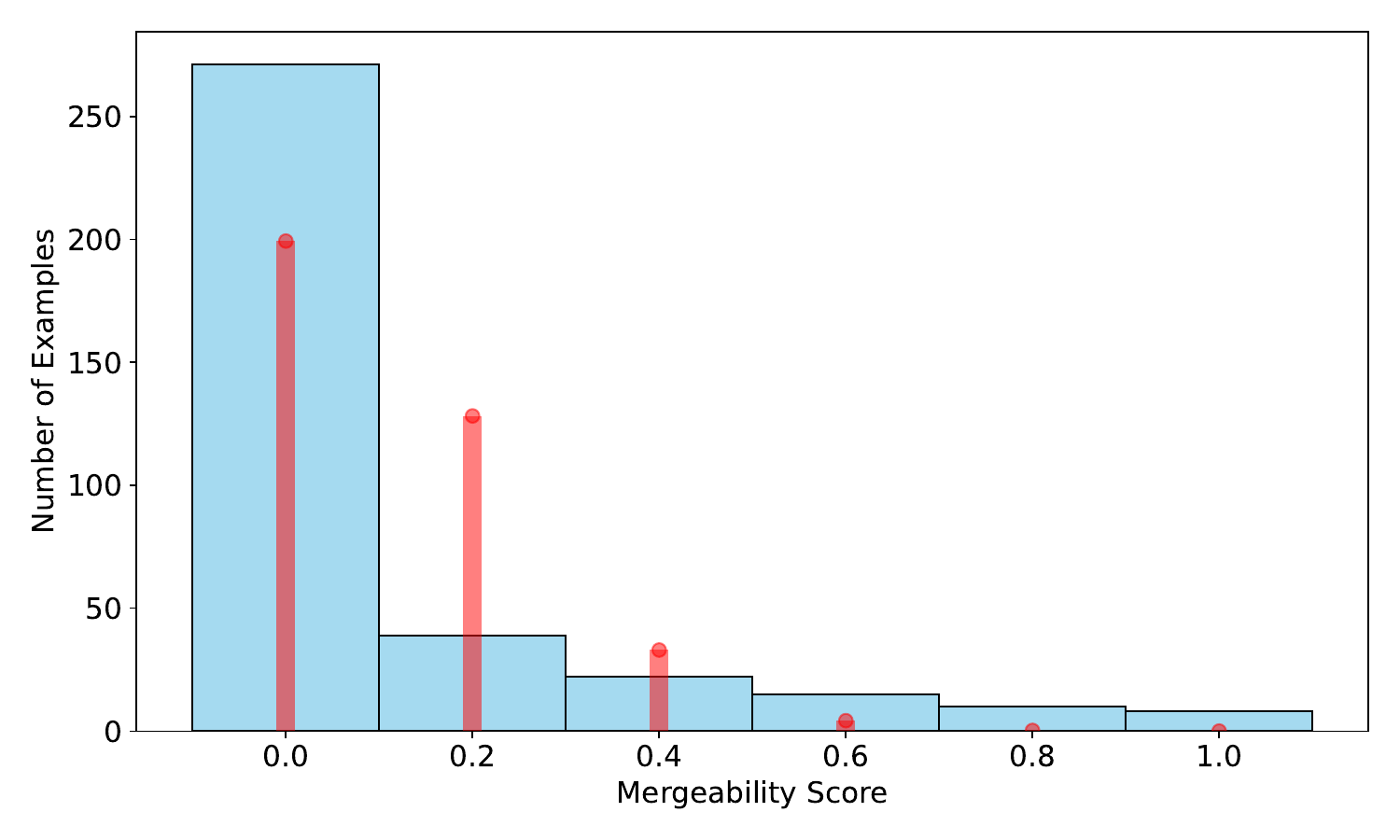}
    \caption{Mergeability scores distribution of Qwen2.5-3B on the PopQA dataset with $N=5, M=50$ and full finetuning training. The scores were calculated using TIES merging algorithm. Blue and wide bars show the mergeability score as empirically calculated. Red and thin bars show the baseline distribution if mergeability was not a model trait, and hence it was a binomial distribution with a fixed success rate.}
    \label{fig:full_ft_ties_histogram}
\end{figure}

\begin{figure}
    \centering
    \includegraphics[width=1\linewidth]{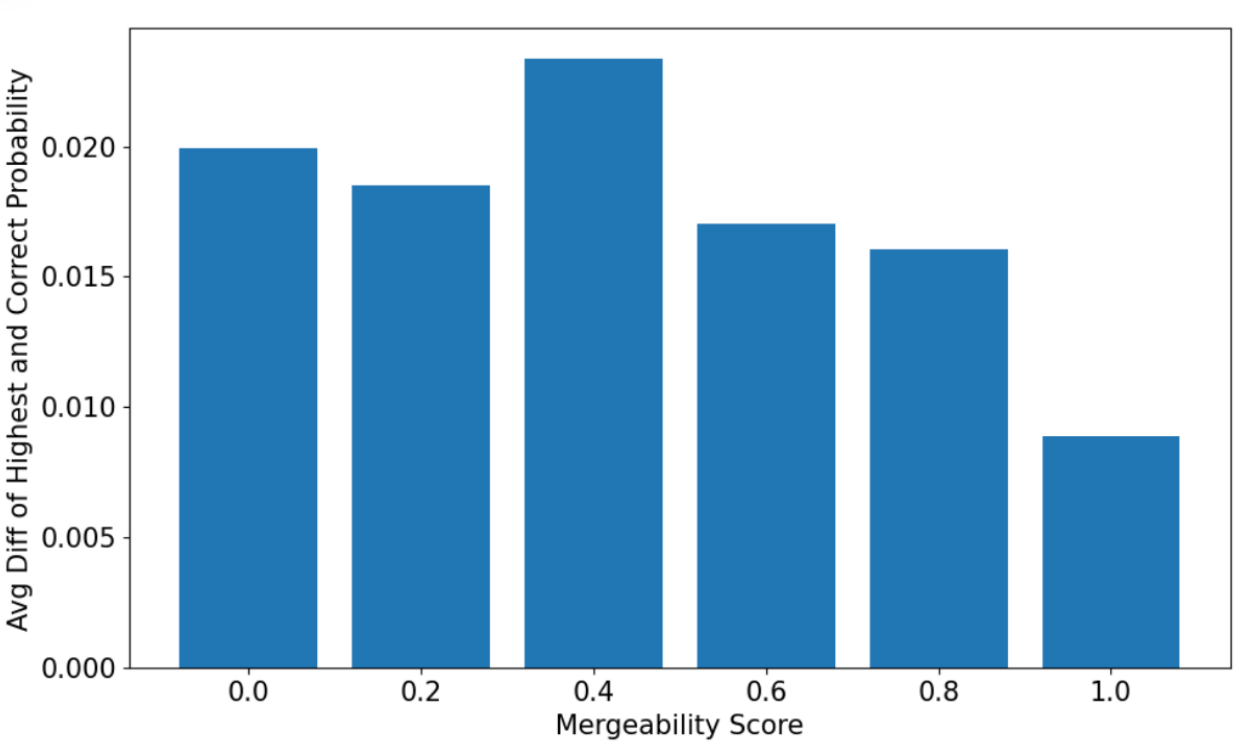}
    \caption{PopQA average difference between the highest and the correct answer probability in the base model (Qwen). Mergeability scores are for examples trained with full finetuning. The scores were calculated using TIES merging. We observe that the gap generally decreases with mergeability, implying that examples with better base model knowledge are more mergeable.}
    \label{fig:full_ft_ties_results}
\end{figure}

To verify that mergeability is not a phenomenon exclusive to parameter-efficient LoRA fine-tuning, we extended our evaluation to full model fine-tuning. We utilized the same Qwen 3B base model and the PopQA dataset experimental setup described in \S\ref{sec:setup}, but trained all model parameters instead of using LoRA adapters. Since the \textsc{Knots} algorithm is designed for LoRA weight merging, we employed two alternative merging algorithms suitable for full weights: TIES-Merging and simple mean merging. Figures \ref{fig:full_ft_mean_histogram} and \ref{fig:full_ft_ties_histogram} show that mergeability also occurs in the full finetuning case. Experimental results in Figures \ref{fig:full_ft_mean_results} and \ref{fig:full_ft_ties_results} generally follow the expected decreasing trend. Those results show that our findings, examples where the base model has better knowledge of are more mergeable, also hold for the full finetuning training.

\subsection{Different Merging Algorithms}
\label{sec:additional_merging_algo}

\begin{figure}[]
    \centering
    \includegraphics[width=1\linewidth]{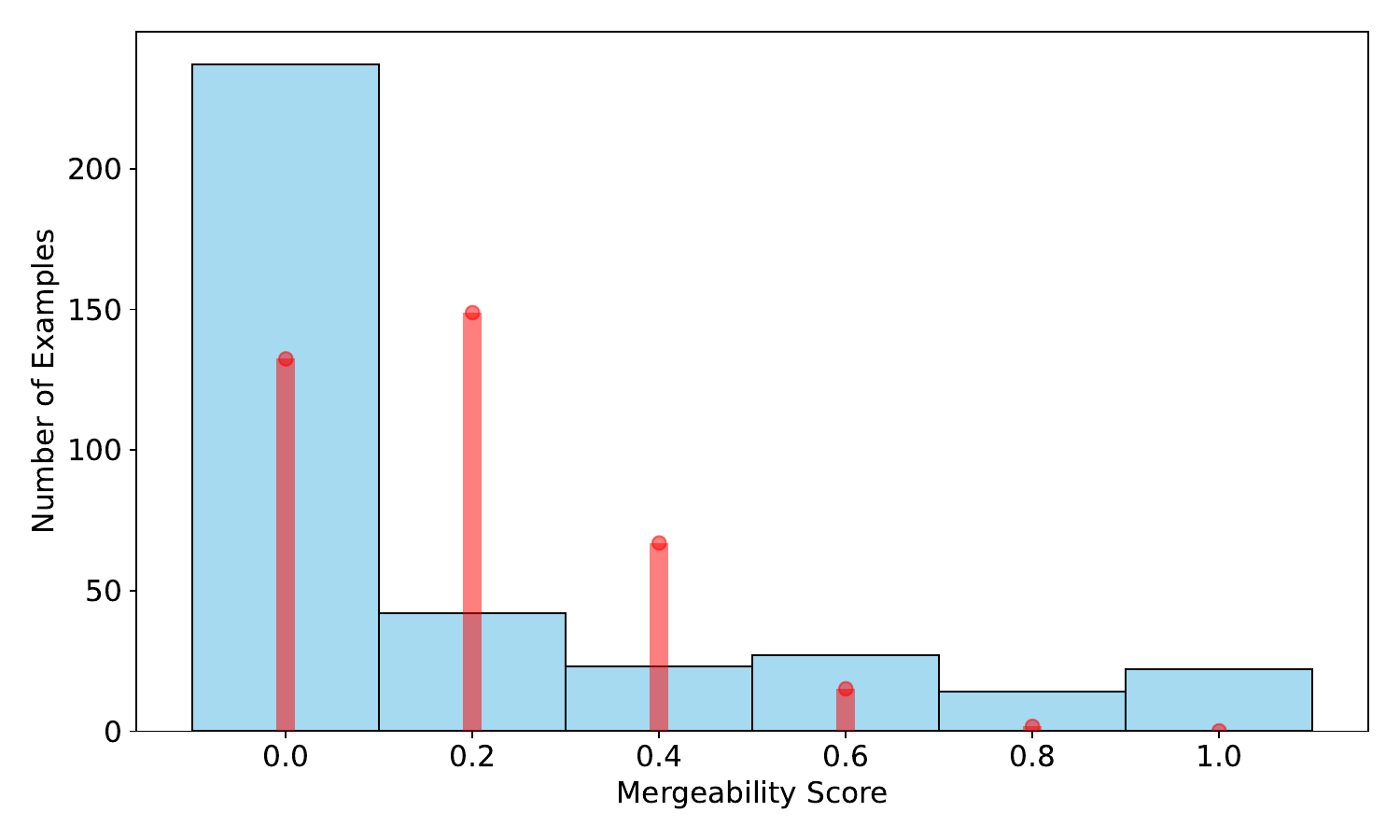}
    \caption{Mergeability scores distribution of Qwen2.5-3B on the PopQA dataset with $N=5, M=50$ and LoRA $r=64$. The scores were calculated using TIES merging algorithm. Blue and wide bars show the mergeability score as empirically calculated. Red and thin bars show the baseline distribution if mergability was not a model trait, and hence it was a binomial distribution with a fixed success rate.}
    \label{fig:qwen_ties_histogram}
\end{figure}

\begin{figure}
    \centering
    \includegraphics[width=1\linewidth]{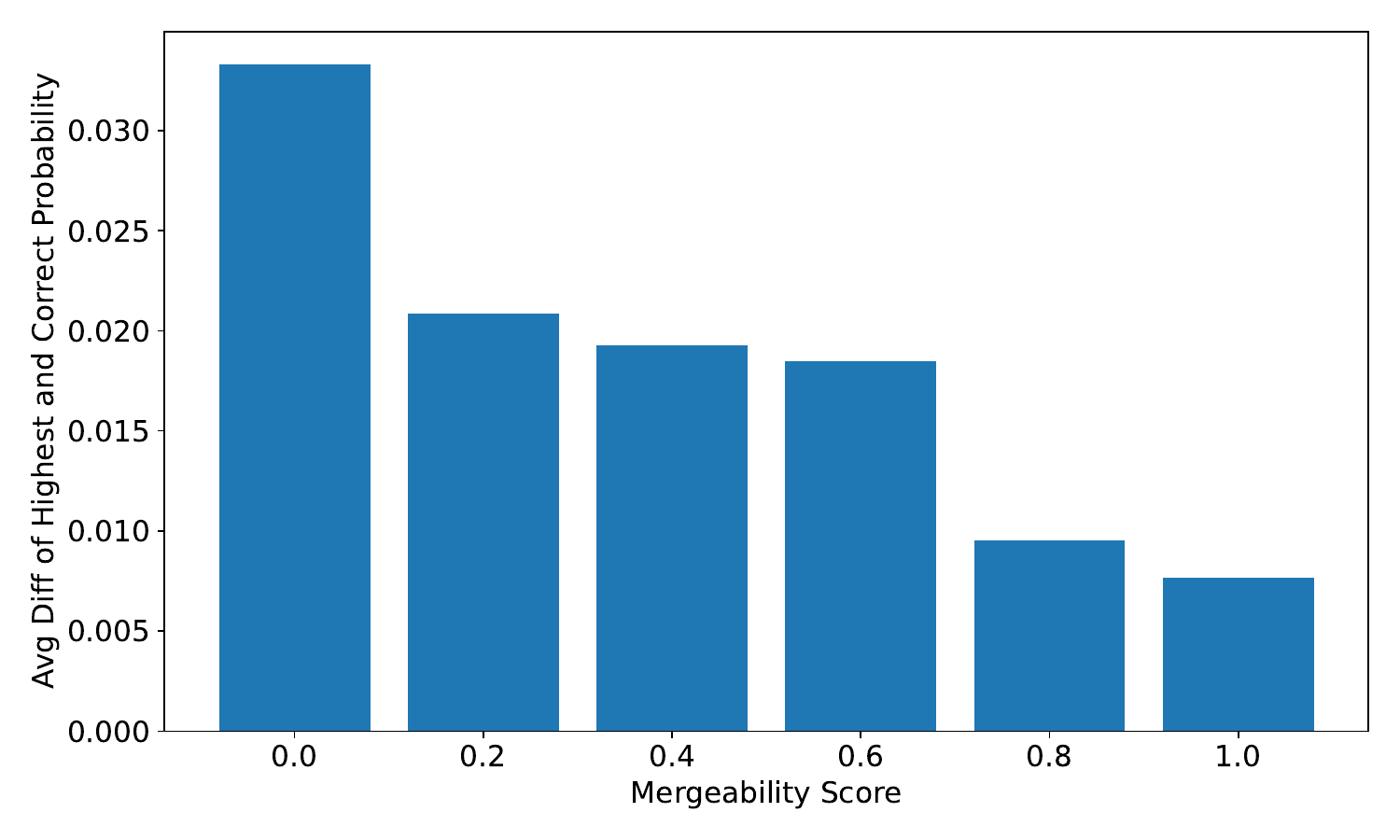}
    \caption{PopQA average difference between the highest and the correct answer probability in the base model (Qwen). The scores were calculated using TIES merging. We observe that the gap generally decreases with mergeability, implying that examples with better base model knowledge are more mergeable.}
    \label{fig:qwen_ties_results}
\end{figure}

\begin{figure}
    \centering
    \includegraphics[width=1\linewidth]{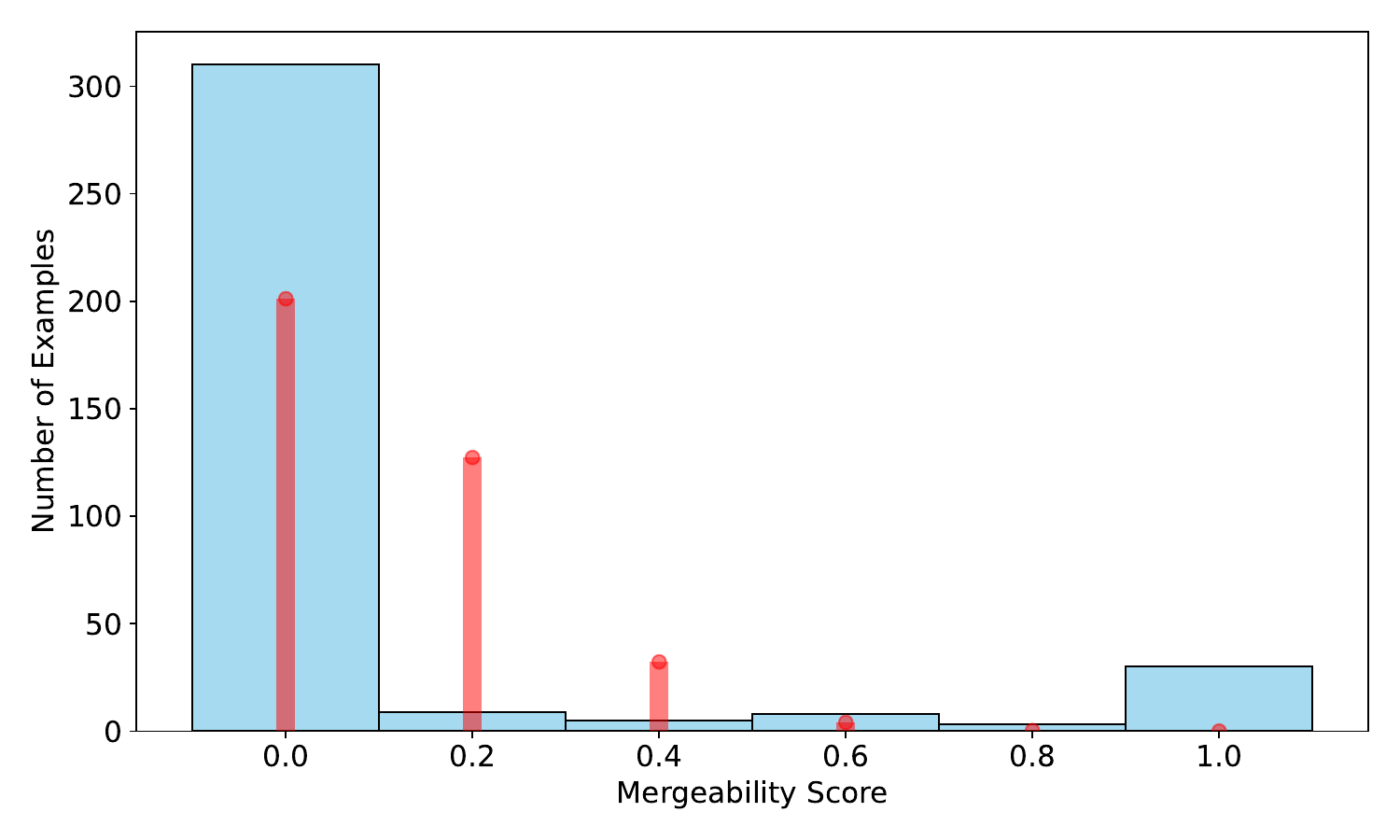}
    \caption{Mergeability scores distribution of Qwen2.5-3B on the PopQA dataset with $N=5, M=50$ and LoRA $r=64$. The scores were calculated using mean merging. Blue and wide bars show the mergeability score as empirically calculated. Red and thin bars show the baseline distribution if mergability was not a model trait, and hence it was a binomial distribution with a fixed success rate.}
    \label{fig:qwen_mean_histogram}
\end{figure}

\begin{figure}
    \centering
    \includegraphics[width=1\linewidth]{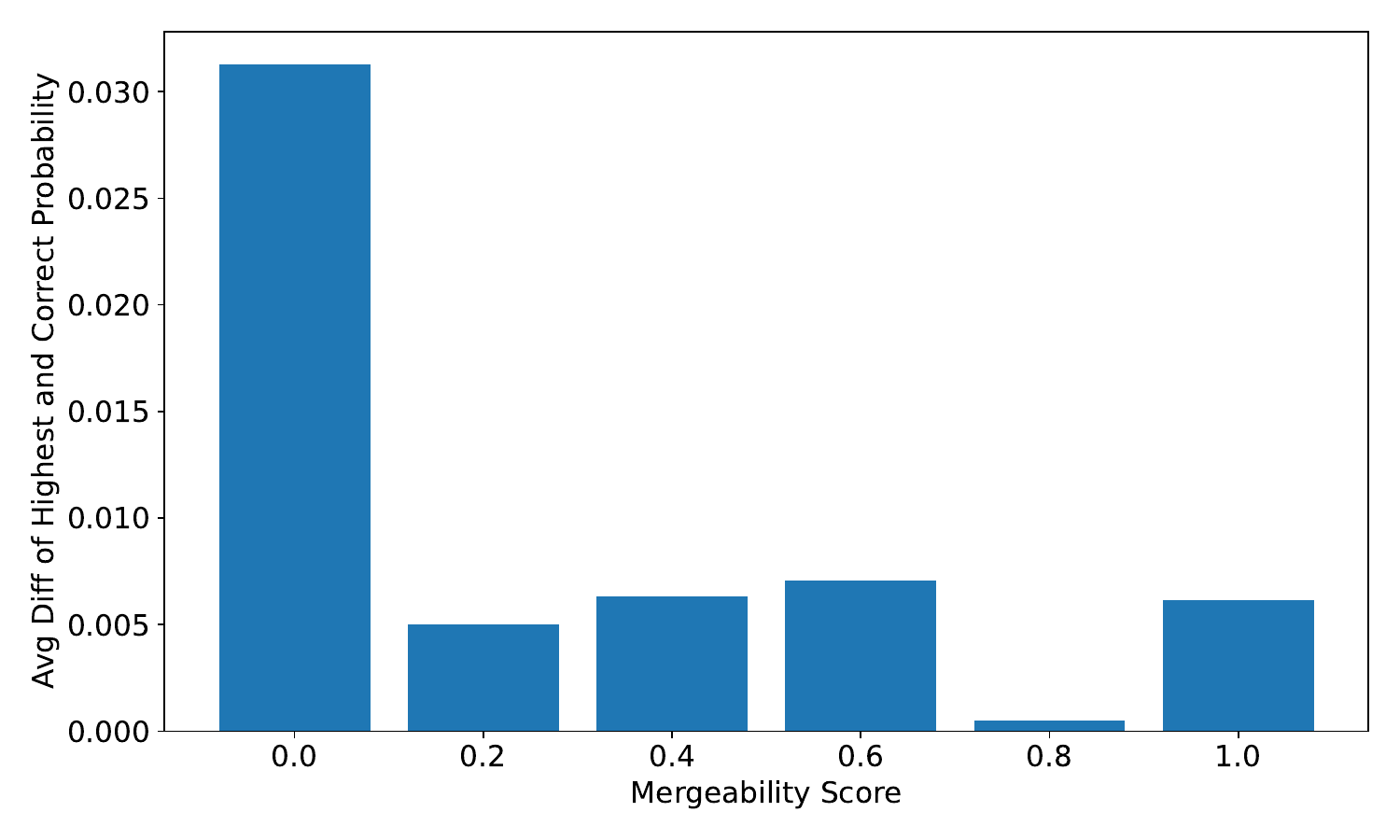}
    \caption{PopQA average difference between the highest and the correct answer probability in the base model (Qwen). The scores were calculated using mean merging. We do not observe a correlation with the mergeability score. This might be exaplined by the low number of examples with $0<S<1$.}
    \label{fig:qwen_mean_results}
\end{figure}

In addition to the main paper results using \textsc{Knots} merging algorithm, we also examined other merging algorithms - TIES and mean. We used the Qwen 3B base model and the PopQA dataset experimental setup described in \S\ref{sec:setup} with the change of using a different merging algorithm for mergeability measurements. Figure \ref{fig:qwen_ties_histogram} and \ref{fig:qwen_mean_histogram} show the mergeability score distribution for TIES and mean merging, respectively. TIES distribution shows similar trends to \textsc{Knots} (Figure \ref{fig:mergeability_score_qwen_5} and supports the existence of mergeability. Mean distribution shows similar trends to Figure \ref{fig:different_merging_algorithms}, with a high number of examples in the $S=1.0$ bin and a very small number of examples in the middle bins ($0<S<1$). As discussed in (\S\ref{sec:merging_affect}), we can attribute this to the lack of any conflict mitigation in mean merging. TIES experimental results (Figure \ref{fig:qwen_ties_results}) show the same decreasing trend between base model knowledge and mergeability score, revealing that our results also hold for TIES merging. When using mean merging, the experimental results (Figure \ref{fig:qwen_mean_results}) does not show the decreasing trend anymore. We can attribute this to the low number of examples with $0<S<1$ that may lead to inaccurate evaluation. 

\begin{figure}[]
    \centering
    \includegraphics[width=1\linewidth]{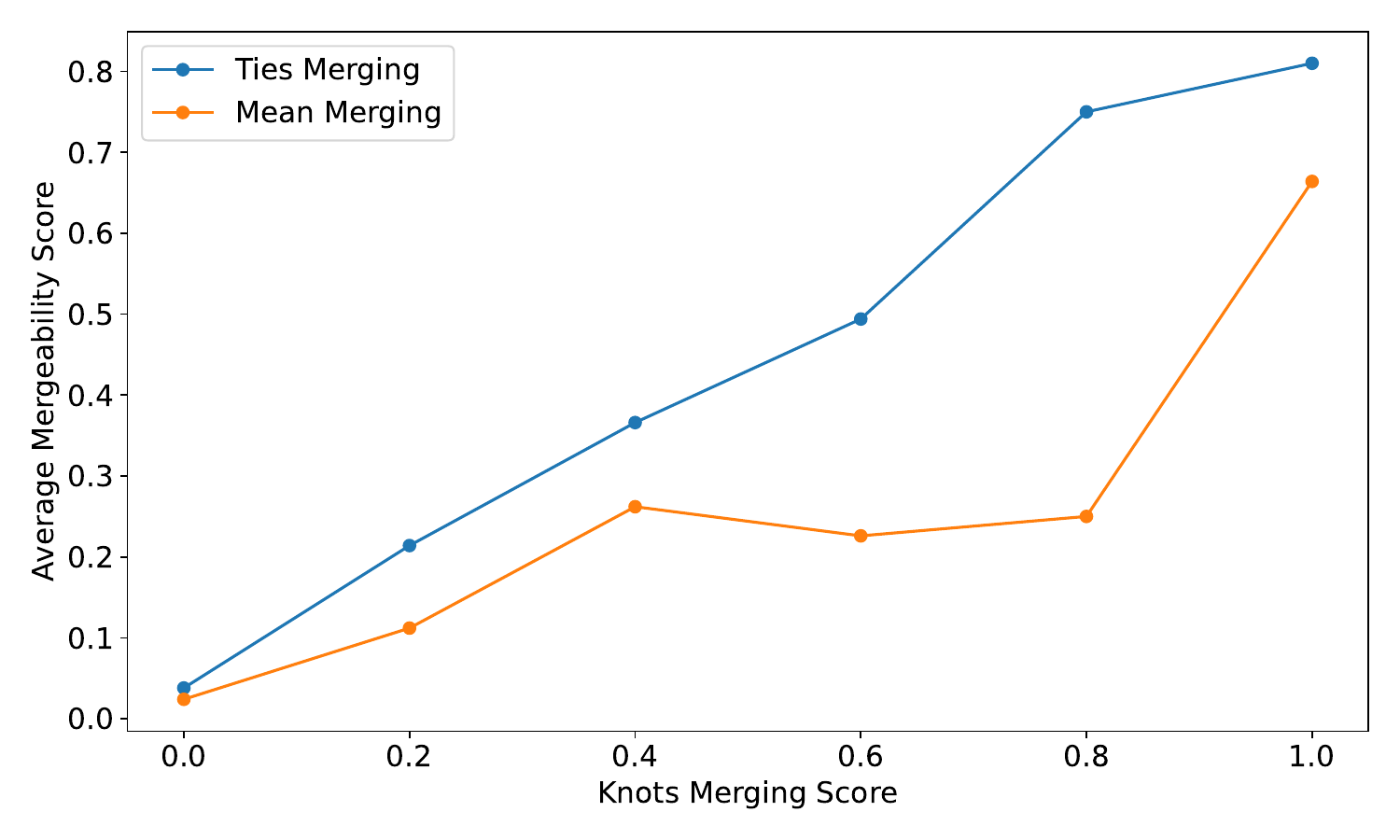}
    \caption{We compare the mergeability score calculated using \textsc{Knots} to other merging algorithms. Results are for Qwen2.5-3B on the PopQA dataset. The x-axis is the mergeability score when calculated using \textsc{Knots}. The y-axis shows the average mergeability score of those examples when calculated with other merging algorithms. We observe an increasing trend for both mean and TIES.}
    \label{fig:knots_vs_ties_vs_mean}
\end{figure}

We further tested how the mergeability score of examples changes as we change the merging algorithm. Figure \ref{fig:knots_vs_ties_vs_mean} shows the mergeability score of examples using TIES or mean merging compared to their score as calculated when using \textsc{Knots} as the merging algorithm. The x-axis is the mergeability
score when using \textsc{Knots} merging. The y-axis
shows the average mergeability score of those examples
when experimenting with different merging algorithms - TIES or mean. We observe an increasing trend for both merging algorithms, which indicates that examples with a higher mergeability score obtained using \textsc{Knots} also had a higher mergeability score when experimented with other merging algorithms. We also see that \textsc{Knots} and TIES scores are more similar compared to mean. This can be explained by the reason that both use a conflict mitigation algorithm compared to mean.

\end{document}